\documentclass{article} 
\usepackage{iclr2026_conference,times}
\iclrfinalcopy

\usepackage{amsmath,amsfonts,bm}

\def\eqref#1{equation~\ref{#1}}

\def\1{\bm{1}}

\usepackage[T1]{fontenc}
\usepackage[utf8]{inputenc}

\usepackage[skins,breakable]{tcolorbox}
\usepackage{fvextra} 

\newcommand{\ShowMD}[2]{%
  \begin{tcolorbox}[breakable,title={#1}]
    \VerbatimInput[
      breaklines=true,    
      breakanywhere=true, 
      obeytabs=true, tabsize=2,
      fontsize=\footnotesize
    ]{#2}%
  \end{tcolorbox}%
}

\usepackage{hyperref}
\usepackage{url}
\usepackage{graphicx}
\usepackage{subcaption}
\usepackage{wrapfig}


\usepackage{tikz}
\usetikzlibrary{arrows.meta,positioning,shadows.blur}
\usepackage[final]{microtype}

\usepackage{tgheros} 

\usepackage{relsize} 

\usepackage{xcolor} 
\definecolor{SecondBest}{HTML}{1A73E8}
\usepackage{listings}
\usepackage{booktabs}  
\usepackage{siunitx}   
\hypersetup{
    colorlinks=true,
    linkcolor=blue,
    citecolor=black,
    filecolor=magenta,
    urlcolor=cyan,
}
\definecolor{FlowFill}{HTML}{FDB241}
\definecolor{FlowStroke}{HTML}{B36A10}
\definecolor{FlowText}{HTML}{000000}
\definecolor{FlowArrow}{HTML}{0F8C8C}

\newlength\FlowOuterW
\newlength\FlowPadX
\newlength\FlowPadY
\providecolor{FlowFill}{rgb}{0.97,0.86,0.64}
\providecolor{FlowStroke}{rgb}{0.65,0.42,0.00}
\providecolor{FlowArrow}{rgb}{0.10,0.55,0.80}

\providecommand{\flowfontsize}{\normalsize}

\tikzset{
  flowboxTight/.style={
    draw=FlowStroke, fill=FlowFill,
    very thick, rounded corners=10pt,
    text width=\dimexpr\FlowOuterW-2\FlowPadX\relax,
    minimum height=2.4cm,        
    inner xsep=\FlowPadX,
    inner ysep=\FlowPadY,
    align=center,
    text=black, text opacity=1,
    font=\normalfont\bfseries\flowfontsize,
    blur shadow={shadow xshift=1.5pt, shadow yshift=-1.5pt,
                 blur radius=2.5pt, opacity=0.14}
  },
  flowarrow/.style={
    -{Stealth[length=4.6mm,width=3.0mm]},
    line width=2pt,
    draw=FlowArrow
  }
}

\title{RefGrader: Automated Grading of Mathematical Competition Proofs using Agentic Workflows}

\author{%
\textbf{Hamed Mahdavi}$^{1}$ \quad \textbf{Pouria Mahdavinia}$^{1}$ \quad \textbf{Samira Malek}$^{1}$ \quad \textbf{Pegah Mohammadipour}$^{1}$ \\
\textbf{Alireza Hashemi}$^{2}$ \quad \textbf{Majid Daliri}$^{3}$ \quad \textbf{Alireza Farhadi}$^{4}$ \quad \textbf{Amir Khasahmadi}$^{5}$ \\
\textbf{Niloofar Mireshghallah}$^{6}$ \quad \textbf{Vasant Honavar}$^{1}$ \\[4mm]
$^{1}$Pennsylvania State University \quad
$^{2}$City University of New York \quad
$^{3}$New York University \\
$^{4}$Amirkabir University of Technology \quad
$^{5}$Autodesk \quad
$^{6}$Carnegie Mellon University
}
\begin{document}

\maketitle

\begin{abstract}
   State-of-the-art (SOTA) LLMs have progressed from struggling on proof-based Olympiad problems to solving most of the IMO 2025 problems, with leading systems reportedly handling 5 of 6 problems. Given this progress, we assess how well these models can grade proofs: detecting errors, judging their severity, and assigning fair scores beyond binary correctness. We study proof-analysis capabilities using a corpus of 90 Gemini 2.5 Pro–generated solutions that we grade on a 1–4 scale with detailed error annotations, and on MathArena solution sets for IMO/USAMO 2025 scored on a 0–7 scale. Our analysis shows that models can reliably flag incorrect (including subtly incorrect) solutions but exhibit calibration gaps in how partial credit is assigned. To address this, we introduce agentic workflows that extract and analyze reference solutions and automatically derive problem‑specific rubrics for a multi‑step grading process. We instantiate and compare different design choices for the grading workflows, and evaluate their trade‑offs. Across our annotated corpus and MathArena, our proposed workflows achieve higher agreement with human grades and more consistent handling of partial credit across metrics. We release all code, data, and prompts/logs to facilitate future research. \href{https://github.com/ref-grader/ref-grader}{https://github.com/ref-grader/ref-grader}
\end{abstract}

\section{Introduction}
Until early 2025, state-of-the-art (SOTA) LLMs often failed to produce correct and sound solutions to Olympiad level problems \citep{petrov2025prooforbluff, mahdavi2025brains}. Industry announcements from Google and OpenAI claimed that the advanced versions of their models could achieve gold medal level performance on the IMO 2025, solving 5 of 6 problems within exam time \citep{LuongLockhart2025GeminiDeepThinkIMOGold,wei-openai-imo-2025-proofs}. Independent reproductions report solving 5 of 6 problems using Gemini 2.5 Pro within an agentic, multi-step workflow \citep{huang2025gemini25imo}. As automated judges, they performed unreliably, often near chance, when asked to distinguish invalid solutions from the correct ones or to apply rubrics consistently \citep{mahdavi2025brains, petrov2025prooforbluff}.

These findings raise concerns about using LLMs for automated proof assessment: if models struggle with basic verification and rubric application, automatic grading may be unreliable. However, the cited studies predate recent model advances. Independent evaluations, such as \citet{balunovic_srimatharena_2025}, report notable improvements in solution correctness and proof quality generated by SOTA (non-agentic) LLMs (e.g., Gemini 2.5 Pro), hinting at their potential improvement in proof verification. Evaluating LLMs' mathematical capabilities via final-answer accuracy has become the de facto standard \citep{cobbe2021gsm8k, hendrycks2021measuringmathematicalproblemsolving, fang2024mathodysseybenchmarkingmathematicalproblemsolving, yue2024harpchallenginghumanannotatedmath}. Going beyond final answers to assess proof quality is substantially more challenging. Formal verification offers a principled solution to validation \citep{zheng2022minif2fcrosssystembenchmarkformal, lin2025goedelproverv2, chen2025seedprover, jiang2024leanreasonerboostingcomplexlogical, ren2025deepseekproverv2advancingformalmathematical}, but faces two practical limitations: limited availability of formal corpora and lower readability for humans. An alternative is to binarize proofs and measure agreement with expert judges \citep{dekoninck2025openproofcorpus, guo2025litmustest}, which improves scalability but ignores the issue of partial credits. We emphasize that \textbf{partial credit assignment} will be an \textbf{increasingly important capability} as we move towards more complex LLM-based proof generation systems.

We constructed a corpus of 90 carefully selected problems from \textbf{complex IMO shortlist problems}, alongside Gemini 2.5 Pro–generated solutions for each problem, graded on a 1–4 scale and annotated with precise error types and locations to act as \textbf{rich ground truth for partial credit assignment.} We also use the data gathered from the MathArena IMO/USAMO 2025 solutions scored 0–7. Using Gemini 2.5 Pro with maximum thinking budget, we first assess single-turn grading by comparing model-assigned scores against human grades. Next, we introduce \textbf{Agentic Workflows} that extract and analyze reference solutions to \textbf{automatically design problem-specific grading rubrics (Ref-Grader)}, and we compare design choices: approachability-based weighting (by “aha moment” difficulty), milestone-based rubrics, their hybrid, and a 3-step reference variant without rubric induction. \textbf{Our workflows substantially improve upon single-turn grading} in partial-credit grading across diverse metrics such as Pearson/Spearman, MAE/RMSE, QWK and AC2. \textbf{We validate robustness through systematic ablations and cross-dataset evaluation}. While our workflows may require greater token consumption (both input and output) and thus incur higher costs, the majority of the workflow steps are cacheable, which helps keep the overall cost low.



\section{Related Work}
\textbf{Proof-evaluation corpora:} Benchmarks assessing proofs include the Open Proof Corpus, which aggregates human and model proofs with binary validity labels and expert annotations \citep{dekoninck2025openproofcorpus}, and LitmusTest, which standardizes pass/fail judgments using expert-designed rubrics \citep{guo2025litmustest}. For competition mathematics, MathArena hosts model-generated solutions for IMO/USAMO-style problems with 0-7 scores and judge rationales \citep{balunovic_srimatharena_2025}. Formal settings emphasize verifiable correctness but face constraints in data availability and coverage \citep{lin2025goedelproverv2, zheng2022minif2fcrosssystembenchmarkformal, chen2025seedprover}.

\textbf{LLM-as-a-grader:} Two directions are prominent: rubric-grounded grading across domains and reliability improvements via calibration or multi-agent designs. Recent work spans diverse applications. In physics education, GPT-4o assigns partial credit using self-consistency and human-in-the-loop triage \citep{PhysRevPhysEducRes.21.010126}. In healthcare, open-ended clinical dialogues are evaluated against physician-written, instance-specific criteria \citep{arora2025healthbench}. For expert long-form tasks, expert-validated rubrics map to checklist items \citep{ruan2025expertlongbench}, while rubric-prompted judge distributions benefit from calibration to human ratings \citep{hashemi-etal-2024-llm}. In education and code assessment, rubric specialization and multi-agent judging improve robustness and interpretability \citep{DBLP:conf/icer/PathakGURGJVMAK25,Chu2025GradeOpt}. Per-problem rubrics have been used to diagnose stepwise skills on word problems \citep{Jin2024LLM_Diagnose_Math_Skills}.

\textbf{LLM-as-a-judge:} Complementary work examines models as evaluators to reduce dependence on human annotations \citep{stephan2024calculationadjudicationexaminingllm, li2024llmsasjudgescomprehensivesurveyllmbased, nasrabadi2024jureejudgessafeguardingllm, ning2024picopeerreviewllms}. Recent methods treat assessment as adaptable and task-aware \citep{tan2024largelanguagemodelsdata, dhurandhar2024rankinglargelanguagemodels} and calibrate reliability against human judgments \citep{kim2024demonstrationadaptivecollaborationlarge, ye2024justiceprejudicequantifyingbiases, liu2025aligninghumanjudgementrole}. General-purpose evaluation resources include UltraFeedback \citep{cui2024ultrafeedbackboostinglanguagemodels}, AlpacaEval \citep{dubois2024lengthcontrolledalpacaevalsimpleway}, Chatbot Arena \citep{chiang2024chatbotarenaopenplatform}, and MT-Bench \citep{zheng2023judgingllmasajudgemtbenchchatbot}. For mathematics specifically, judge benchmarks include REASONEVAL \citep{xia2025evaluatingmathematicalreasoningaccuracy}, MATHCHECK \citep{zhou2024modelreallygoodmath}, and SMART-840 \citep{cherian2024evaluatinglargevisionandlanguagemodels}.

\textbf{Reasoning Benchmarks:} A range of datasets evaluate mathematical reasoning in large language models (LLMs) \citep{ahn2024largelanguagemodelsmathematical}, spanning arithmetic-only benchmarks \citep{yuan2023largelanguagemodelsperform} and math word problem (MWP) datasets like GSM8K \citep{cobbe2021gsm8k} and MathQA \citep{amini2019mathqainterpretablemathword} that require logical reasoning \citep{wei2022chainofthought}; related robustness and compositionality benchmarks include GSM1K, Compositional GSM, and Functional MATH \citep{zhang2024gsm1k, hosseini2024compositionalgsm, srivastava2024functional}, while automated theorem proving (ATP) datasets target formal theorem proving \citep{zheng2022minif2fcrosssystembenchmarkformal, yu2024metamathbootstrapmathematicalquestions, jiang2024leanreasonerboostingcomplexlogical}; advanced and Olympiad-level evaluations include CONIC10K, GHOSTS, miniGHOSTS, CHAMP, OlympiadBench, MathOdyssey, and Omni-MATH \citep{wu2023conic10kchallengingmathproblem, frieder2023mathematicalcapabilitieschatgpt, mao-etal-2024-champ, he2024olympiadbench, fang2024mathodysseybenchmarkingmathematicalproblemsolving, gao2024omnimath}, and complementary resources include HARP and NuminaMath \citep{yue2024harpchallenginghumanannotatedmath, numina_math_datasets}.

\textbf{Mathematical Reasoning in LLMs:} Reasoning can be elicited through prompting and inference-time strategies, including Chain-of-Thought and self-consistency \citep{chen2024languagemodelshiddenreasoners, wei2022chainofthought, kojima2023largelanguagemodelszeroshot, havrilla2024glorewhenwhereimprove, wang2023selfconsistencyimproveschainthought, wang2024chainofthoughtreasoningprompting}. Controlled benchmarks reveal gaps between pattern matching and formal reasoning \citep{hendrycks2021measuringmathematicalproblemsolving, mirzadeh2024gsmsymbolicunderstandinglimitationsmathematical}. Complementary work explores reward modeling, self-refinement, and algorithmic decomposition \citep{huang2024largelanguagemodelsselfcorrect, zelikman2023parselalgorithmicreasoninglanguage}.

\section{Datasets}

\subsection{IMO Shortlist Data}

\noindent\textbf{Data Collection.}
We selected 90 challenging problems from the IMO Shortlist dataset (2017-2023). We used a standardized prompt requesting a rigorous solution to each Olympiad-level problem and generated one solution per problem with Gemini 2.5 Pro. The prompt is provided in Appendix \ref{app:solver-prompt}. We then annotated the solutions using the fallacy categories from \citep{mahdavi2025brains}. The list of fallacies is as follows:

\begin{itemize}

\item \textbf{Proof by Example}  
\item \textbf{Proposal Without Verification}  
\item \textbf{Inventing Wrong Facts}  
\item \textbf{Begging the Question (Circular Reasoning)}  
\item \textbf{Solution by Trial-and-Error}  
\item \textbf{Calculation Mistakes}  

\end{itemize}

We adopt the definitions provided in the original paper \citep{mahdavi2025brains}. We additionally introduce a general category, \textbf{Wrong Logical Conclusion}, to tag mathematical errors that do not fit any of the other categories. Evaluators carefully reviewed each solution and annotated each error type and the approximate error location using the following syntax (markup used in the released dataset):

\begin{center}
   \color{blue}
   \texttt{<span class="[Fallacy Type]$^+$"> [Fallacious Statement] </span>}\color{black}
\end{center}

For example, if a fallacy is identified in a generated proof, evaluators mark it as follows:

\color{blue}
\texttt{<span class= "proof-by-example"> As the statement is true for $n=1,2,3$ it is highly probable that it is also true </span>}
\color{black}

When applying fallacy labels, if multiple fallacies fit a given error, we prioritized the most relevant one. When distinct errors co-occurred, we applied multiple fallacy labels. We graded solutions using the following 4-point scale.

\begin{itemize}
\item \textbf{1: Incorrect:} The solution does not contain useful non-trivial information. It contains only incorrect information or restates straightforward facts from the problem. Equivalent to 0/7 or 1/7 in Olympiad grading.
\item \textbf{2: Some Correct Information:} The solution contains a few non-trivial facts derived with some effort but lacks a coherent proof. Equivalent to 2/7 or 3/7 in Olympiad grading. 
\item \textbf{3: Almost Correct:} The solution proves non-trivial parts of the argument but omits one non-trivial part of the proof. Equivalent to 4/7 or 5/7 in Olympiad grading.
\item \textbf{4: Correct:} The solution proves all required facts and statements
\end{itemize}

We did not adopt the 0-7 Olympiad scale due to the higher probability of inconsistencies between human evaluators, and considerably more human effort to grade at that granularity. Finally, after annotating errors and assigning grades, evaluators provided a brief explanation of any issues in a dataset field labeled "Final Comment". 

\begin{figure*}[t]
   \centering
   \begin{subfigure}[t]{0.32\textwidth}
       \centering
       \includegraphics[width=\linewidth]{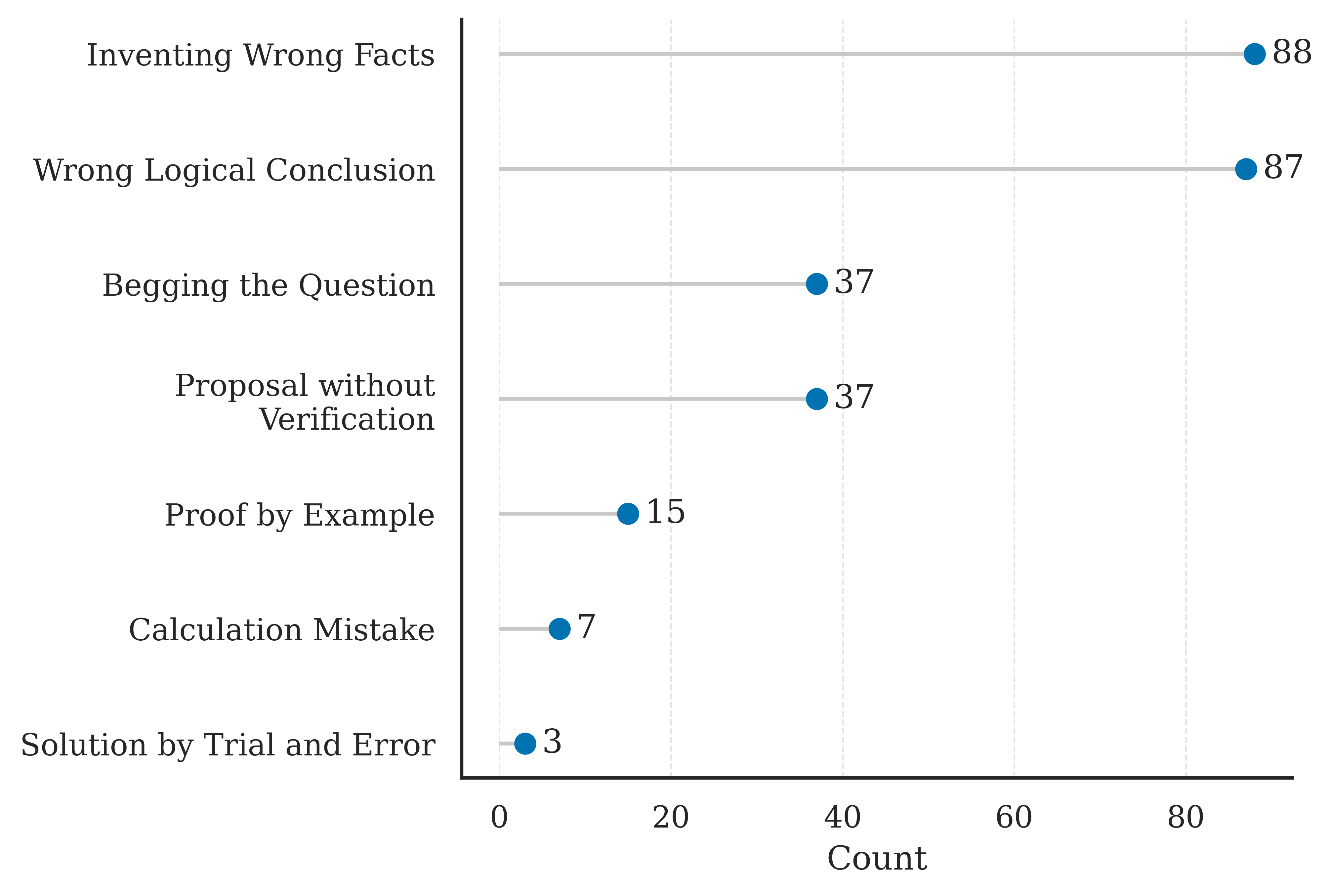}
       \caption{Error frequencies by fallacy category.}
       \label{fig:error-frequencies}
   \end{subfigure}\hfill
   \begin{subfigure}[t]{0.32\textwidth}
       \centering
       \includegraphics[width=\linewidth]{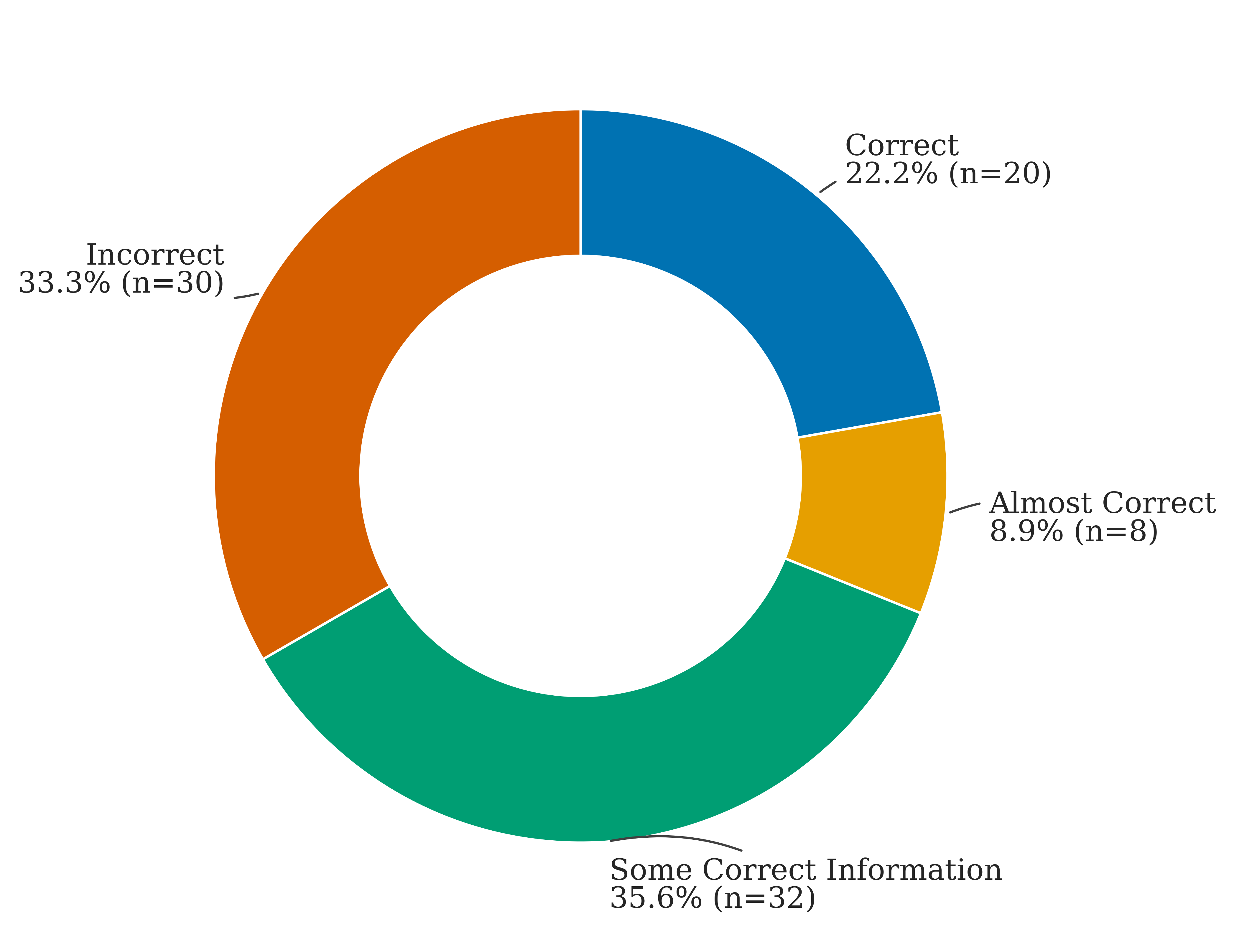}
       \caption{Distribution of solution labels (percentages and counts).}
       \label{fig:labels-distribution}
   \end{subfigure}\hfill
   \begin{subfigure}[t]{0.32\textwidth}
       \centering
       \includegraphics[width=\linewidth]{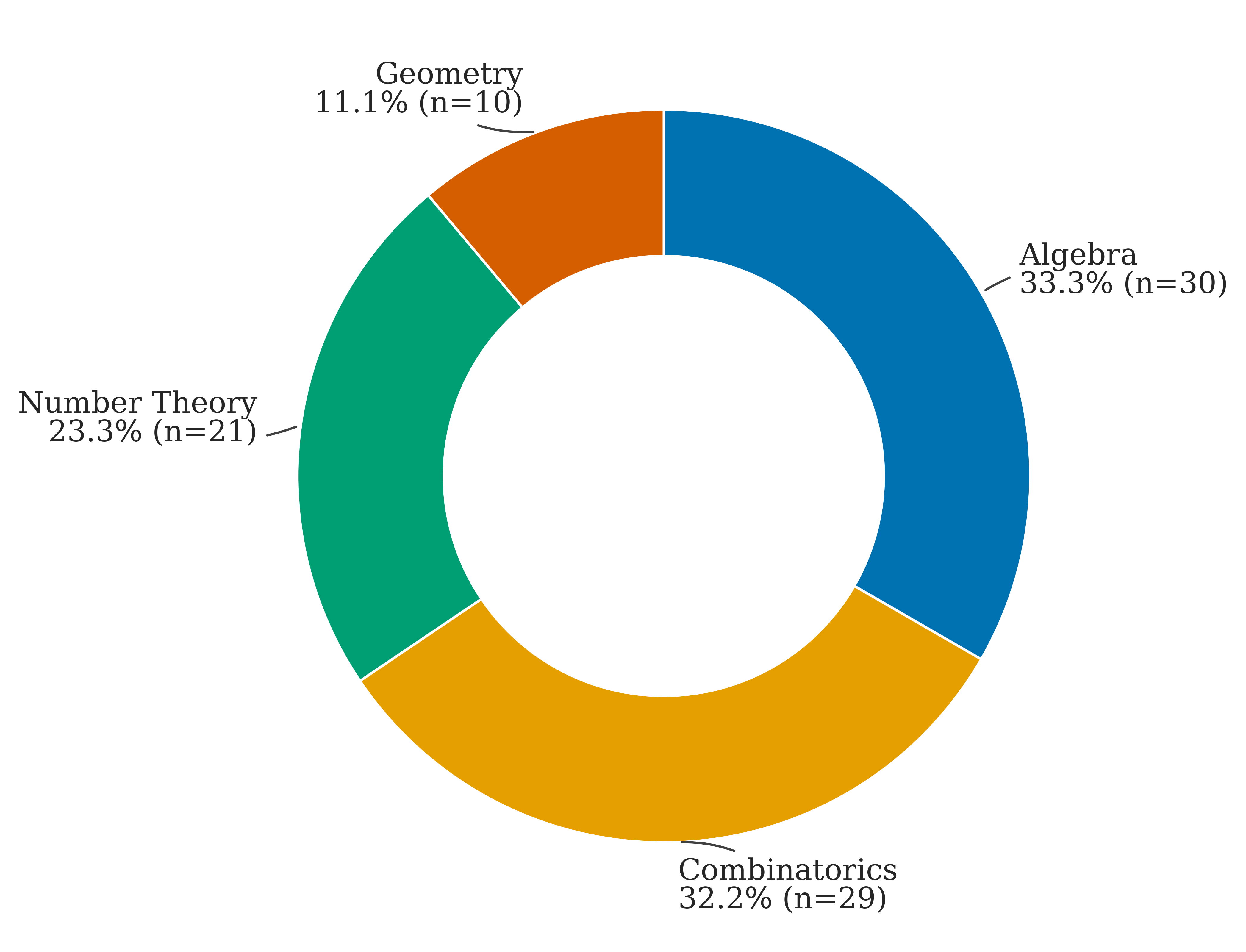}
       \caption{Problem topics (percentages and counts).}
       \label{fig:topic-distribution}
   \end{subfigure}
   \caption{Dataset summaries and error analysis for the IMO Shortlist dataset}
   \label{fig:combined-dataset-summary}
\end{figure*}

\noindent\textbf{Dataset Statistics.}
Figures~\ref{fig:error-frequencies},~\ref{fig:labels-distribution} and \ref{fig:topic-distribution} summarize dataset statistics: error frequencies by fallacy category, the distribution of solution labels, and the topical composition of problems. Relative to the models analyzed by \citet{mahdavi2025brains}, Gemini 2.5 Pro yields a smaller share of incorrect solutions (Fig.~\ref{fig:labels-distribution}) and fewer naive errors (e.g., Proof by Example, Solution by Trial-and-Error; Fig.~\ref{fig:error-frequencies}).

\subsection{MathArena Data}

\begin{wrapfigure}{r}{0.40\textwidth}
   \vspace{-6mm}
   \centering
   \includegraphics[width=0.40\textwidth]{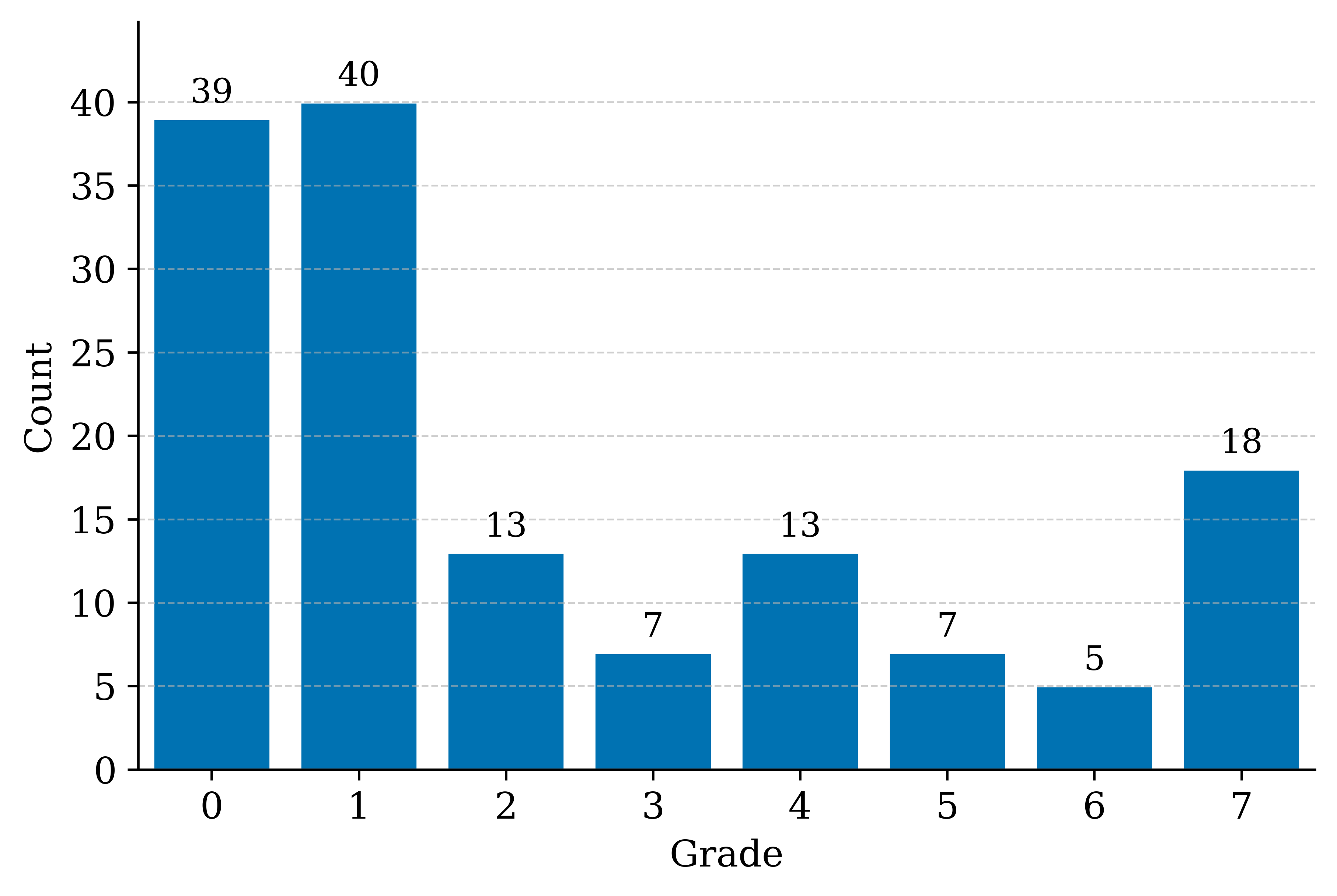}
\caption{Grade distribution for the MathArena dataset}
   \label{fig:grades-hist}
   \vspace{-4mm}
\end{wrapfigure}

We collected 385 solutions for IMO and USAMO 2025 from the MathArena website. The solutions were generated by the following models: Grok 3 (Think), DeepSeek\textendash R1\textendash 0528, Gemini 2.5 Pro, Gemini 2.0 Flash Thinking, QwQ\textendash 32B, DeepSeek\textendash R1, o1\textendash pro (high), o3\textendash mini (high), o4\textendash mini (high), Grok 4, o3 (high), and Claude\textendash 3.7\textendash Sonnet (Think). MathArena conducts independent evaluations of model performance on contest-level problems. Solutions are graded by human judges on a 0-7 scale. The MathArena grade distribution is zero-inflated because many model-generated solutions receive a zero on these challenging problems. To balance the dataset for analysis and visualization, we subsampled zero-scores with probability 0.14 (applying this subsample consistently in the figures and tables for this section). Figure \ref{fig:grades-hist} shows the resulting grade distribution.

\section{Evaluation Setting}
Our goal is to evaluate LLMs as graders of mathematical proofs on the IMO Shortlist and MathArena datasets. Let $\mathcal{D}=\{(p_i,s_i)\}_{i=1}^n$ denote problem–solution pairs with associated ground-truth grades $\{g_i\}_{i=1}^n$. For each instance $i$, let $R_i = \{ r_{ij} \}_{j=1}^{m_i}$ denote the set of correct reference solutions. The grading procedure (agentic workflow) takes $(p_i,s_i,R_i)$ as input and outputs a predicted grade $\hat{g}_i$. For all experiments, the end result is an LLM output in a structured format that includes the predicted grade $\hat{g}_i$ and, when available, step-by-step analysis, identified errors, clarity/structure/notation tags, and constructive feedback.

To assess agreement between $\{\hat{g}_i\}$ and $\{g_i\}$, we report Pearson and Spearman correlations, mean absolute error (MAE), root mean squared error (RMSE), off-by-one and off-by-two tolerance rates, quadratic weighted kappa (QWK), and Gwet's AC2.

The first four metrics are well-known and we omit their definitions here. We define the off-by-one and off-by-two metrics as:
$$
\textbf{Off-by-one} = \frac{1}{n}\sum_{i=1}^{n}\mathbf{1}\{|g_i - \hat{g}_i|\le 1\},\quad
\textbf{Off-by-two} = \frac{1}{n}\sum_{i=1}^{n}\mathbf{1}\{|g_i - \hat{g}_i|\le 2\}.
$$
These summarize near-miss accuracy when small deviations are acceptable but large errors are costly.

\paragraph{Quadratic weighted kappa (QWK).}
QWK \citep{cohen1968weighted} measures agreement on ordinal labels while accounting for chance. With $K$ grade categories, let $O,E\in\mathbb{R}^{K\times K}$ be the observed and expected confusion matrices, and let $w_{ij}={(i-j)^2}/{(K-1)^2}$. Then
$$
\kappa = 1 - \frac{\sum_{i,j} w_{ij} O_{ij}}{\sum_{i,j} w_{ij} E_{ij}}.
$$
Under rater independence, the expected matrix $E$ uses the raters' marginal grade distributions $p$ and $q$, with entries $E_{ij}=p_i q_j$ (or $n p_i q_j$ for counts), where $p_i=\sum_{j} O_{ij}$ and $q_j=\sum_{i} O_{ij}$. This sets the chance baseline. When marginals are skewed, the baseline is high and QWK can be low even if raw agreement is high.

\paragraph{Gwet's AC2.}
AC2 \citep{gwet2014handbook} uses the same ordinal weights but a chance model that is less sensitive to skew. It replaces the independence baseline $p_i q_j$ with a pooled marginal distribution $\pi_i$ computed across raters (for two raters with marginals $p$ and $q$, $\pi_i=(p_i+q_i)/2$, or $(n_i^A+n_i^B)/(2n)$ with counts), and computes expected weighted disagreement $D_e=\sum_{i,j} w_{ij}\,\pi_i\,\pi_j$. With observed weighted disagreement $D_o=\sum_{i,j} w_{ij} P_{ij}$ for the normalized table $P$, 
$$
\mathrm{AC2} = 1 - \frac{D_o}{D_e}.
$$
Using pooled marginals reduces the impact of imbalanced category frequencies, making AC2 typically more stable than QWK when categories are skewed.

Because grades are discrete and skewed, \textbf{no single metric is sufficiently reliable on its own}, therefore, we draw conclusions from a comprehensive analysis across all metrics. Pearson and Spearman track association but not error cost, and both can look strong when a model predicts the most frequent grade. Ties on a coarse 0 to 7 scale further reduce Spearman's resolution. MAE is interpretable in points and RMSE highlights large mistakes, yet either can look good if the model mostly predicts the same common grade. Off-by-one and off-by-two quantify near-miss tolerance but can be inflated by always predicting central or majority grades. QWK is ordinal and chance-corrected, but it is sensitive to skewed marginals and can show a paradox: raw agreement can be high yet QWK low when grade frequencies are highly imbalanced, because the chance baseline is also high. AC2 keeps ordinal weights but uses pooled marginals, making it more robust to imbalance. It is typically more stable than QWK when categories are skewed, though results still depend on the weight choice and very rare grades. Since no metric is perfect, we report them together to get a balanced view of ordering, error size, near-miss tolerance, and agreement beyond chance in this imbalanced setting.

For the IMO Shortlist, we map the 4-point scale to the 0-7 scale using $m(x)=2x-1$ for $x\in\{1,2,3,4\}$. MathArena is already on the 0-7 scale.

\section{Experimental Results}
We first evaluate the performance of LLMs for single-turn proof grading and present quantitative metrics alongside qualitative visualizations.

\subsection{Single-turn Grading}

In our first experiment, we focus on evaluating the performance of LLMs on grading proofs in a single-turn setting. We add the problem and solution in the context and ask the LLM to analyze the proof step-by-step, find all of its errors, and then grade the proof on a 0-7 scale. We use the following definition for the grading scale:
\noindent\begin{center}
\footnotesize
\begin{tabular}{lc@{\hspace{1em}}lc}
\toprule
\textbf{Definition} & \textbf{Score} & \textbf{Definition} & \textbf{Score} \\
\midrule
No progress & 0 & Substantial progress & 4 \\
Understaning trace & 1 & One small flaw & 5 \\
Minor progress & 2 & Negligible issues & 6 \\
Partial progress & 3 & Perfect & 7 \\
\bottomrule
\end{tabular}
\end{center}

  The full grading prompt used in this setting is provided in Appendix \ref{app:absolute-grader}. The results for MathArena and the IMO Shortlist dataset are shown in Table~\ref{tab:single_turn_grading}. For reference, the \textbf{standard deviation} and \textbf{mean absolute deviation} of real scores are \textbf{2.42} and \textbf{1.87} for MathArena, and \textbf{2.22} and \textbf{1.68} for the IMO Shortlist. If we always predict the majority grade, \textbf{Off-by-one} and \textbf{Off-by-two} are 0.56 and 0.65 on MathArena, and 0.31 and 0.68 on the IMO Shortlist, respectively. Hence, although the correlation metrics indicate non-random association, single-turn grader struggles to predict the real scores accurately.

\begin{figure*}[t]
      \centering
      \begin{subfigure}[t]{0.48\textwidth}
          \centering
          \includegraphics[width=\linewidth]{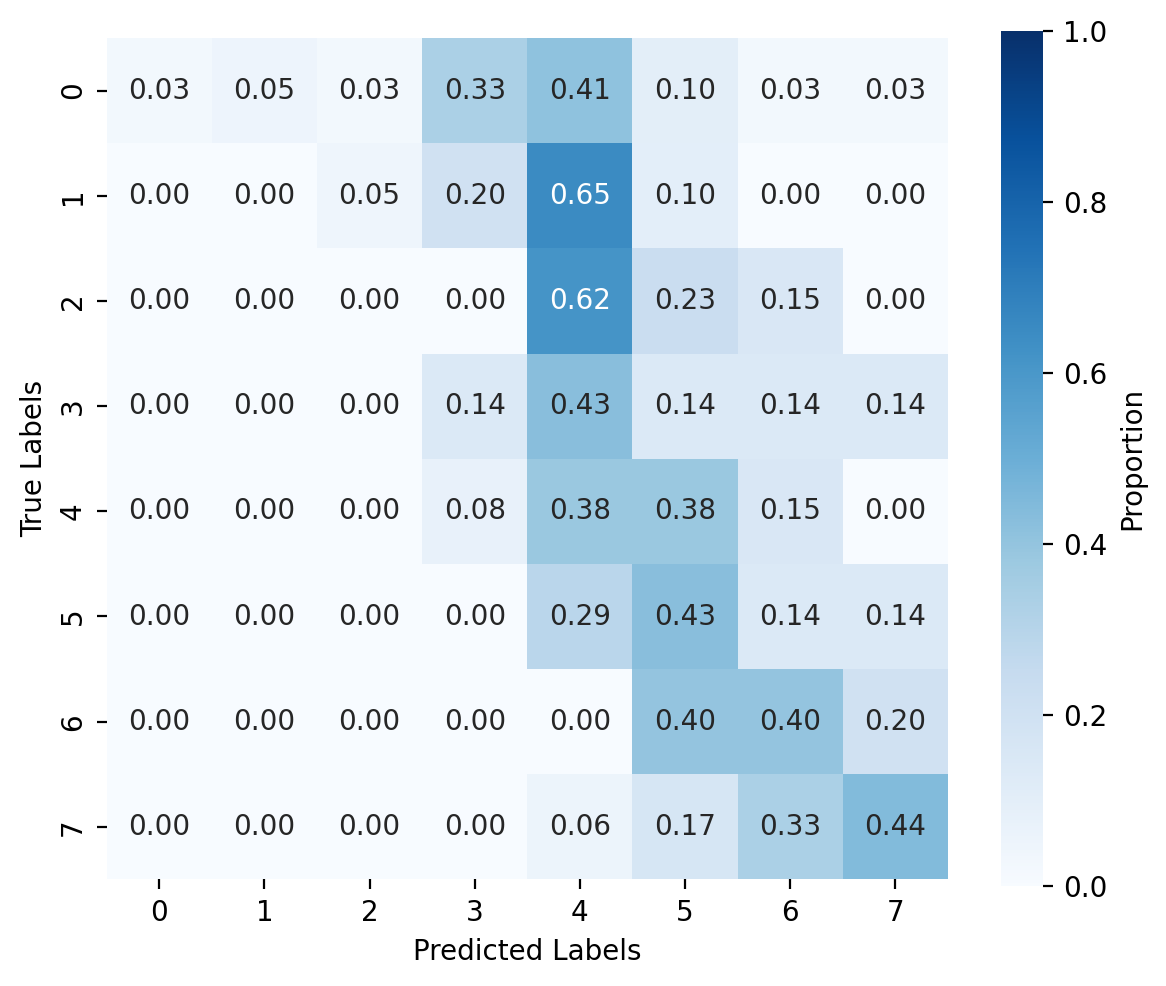}
          \caption{MathArena: normalized confusion matrix.}
          \label{fig:cm-matharena}
      \end{subfigure}\hfill
      \begin{subfigure}[t]{0.48\textwidth}
          \centering
          \includegraphics[width=\linewidth]{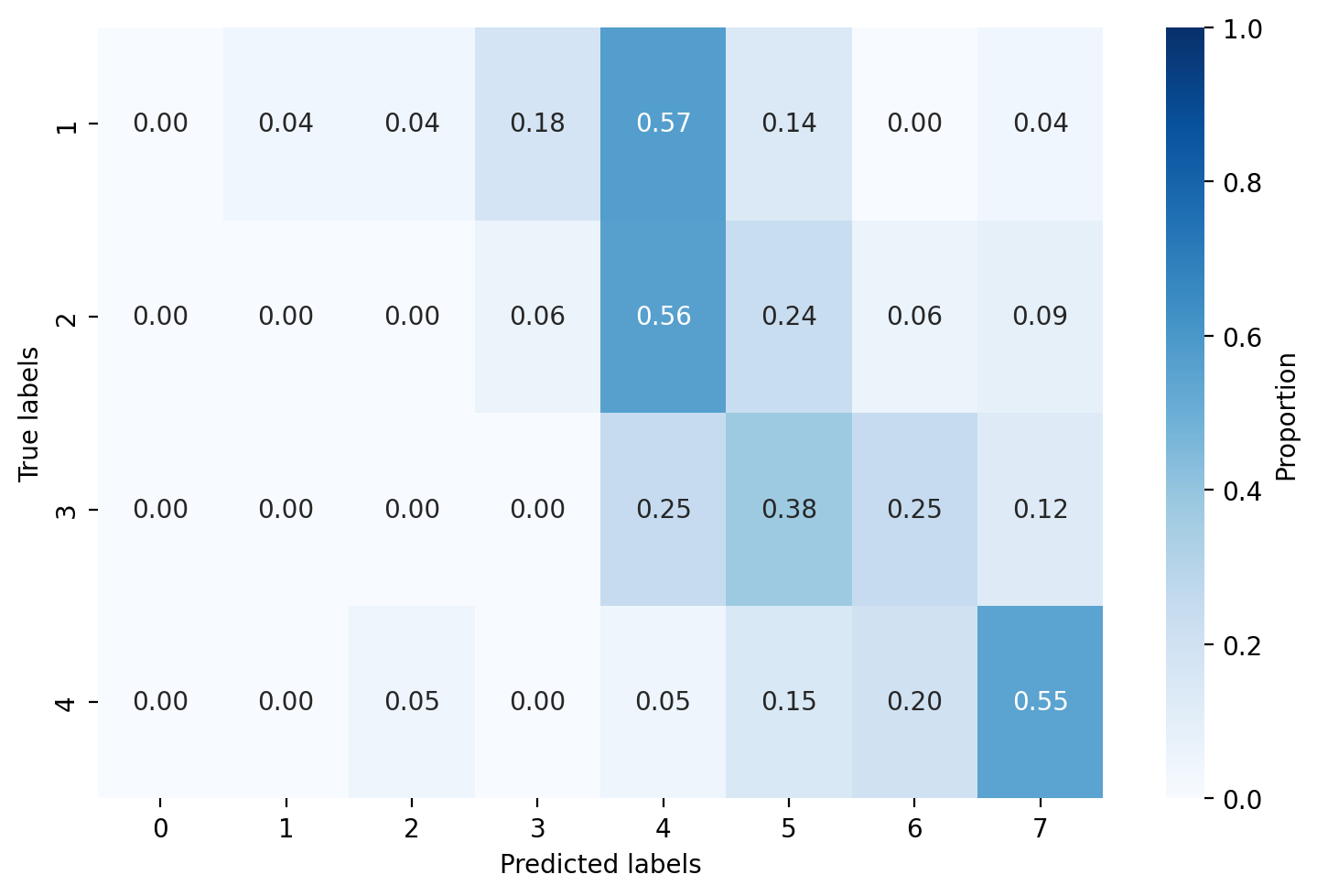}
          \caption{IMO Shortlist: normalized confusion matrix.}
          \label{fig:cm-imo}
      \end{subfigure}
      \vspace{2mm}
      \caption{Normalized confusion matrices for single-turn grading on MathArena and IMO Shortlist.}
  \end{figure*}

  Figures~\ref{fig:cm-matharena} and~\ref{fig:cm-imo} show normalized confusion matrices. On both datasets, the grader tends to over-score very low-grade solutions and partially correct ones (grades 0-4), shifting probability mass to the right of the diagonal. By contrast, solutions with grades $\geq 5$ show a stronger diagonal. This pattern is consistent with the findings of \citet{dekoninck2025openproofcorpus} and \citet{guo2025litmustest}. Under a binarized evaluation (correct vs not correct), performance would be high. More specifically, most off-diagonal mass concentrates a few bins to the right of the true grade for true grades 0-3, indicating an optimistic bias and a tendency to credit incomplete outlines. Misclassifications are predominantly adjacent, and there are fewer solutions with grades 2-6 in both datasets due to data imbalance, which yields higher rank-based measures (Pearson/Spearman) while increasing absolute error (MAE/RMSE). In the bottom-right region, over-scoring is limited, yielding a clearer diagonal and explaining the strong binary separation at threshold 5.

  Conceptually, binary grading is simpler: a reliable verifier can confirm the correctness of a complete solution and identify shortcomings in incomplete or incorrect solutions. However, assessing progress in incomplete solutions is more challenging. Assigning fair partial credit is ambiguous when the model cannot solve the problem using the solution's attempted approach, even if it can solve the problem by a different approach, that alone is not sufficient. We show empirically that leveraging \textbf{a set of candidate reference solutions} within a multi-step grading workflow yields substantially better performance.

\begin{table}[ht]
   \centering
   \small
   \setlength{\tabcolsep}{6pt}  
   \begin{tabular}{l
                  S[table-format=1.3]
                  S[table-format=1.3]
                  S[table-format=1.3]
                   S[table-format=1.3]
                   S[table-format=1.3]
                   S[table-format=1.3]
                   S[table-format=1.3]
                   S[table-format=1.3]}
   \toprule
   \multicolumn{1}{c}{\textbf{Dataset}} &
  \multicolumn{1}{c}{\textbf{Pearson\,\(\uparrow\)}} &
  \multicolumn{1}{c}{\textbf{Spearman\,\(\uparrow\)}} &
  \multicolumn{1}{c}{\textbf{MAE\,\(\downarrow\)}} &
  \multicolumn{1}{c}{\textbf{RMSE\,\(\downarrow\)}} &
  \multicolumn{1}{c}{\textbf{QWK\,\(\uparrow\)}} &
  \multicolumn{1}{c}{\textbf{Off1\,\(\uparrow\)}} &
  \multicolumn{1}{c}{\textbf{Off2\,\(\uparrow\)}} &
  \multicolumn{1}{c}{\textbf{AC2\,\(\uparrow\)}} \\
   \midrule
  Math-Arena    & 0.665 & 0.633 & 2.324 & 2.745 & 0.359 & 0.317 & 0.486 & 0.357 \\
  IMO Shortlist & 0.601 & 0.596 & 1.756 & 2.211 & 0.427 & 0.500 & 0.689 & 0.479 \\
   \bottomrule
   \end{tabular}
  \caption{Single-turn grading results. Higher is better for correlations, QWK, Off1/Off2/AC2. Lower is better for MAE/RMSE.}
    \label{tab:single_turn_grading}
   \end{table}

\subsection{Multi-step Grading with Reference Solutions}
 We next evaluate reference-aided, multi-step grading workflows and ablations. To address the conceptual issue discussed above, we introduce a multi-step reference grading workflow (\emph{Ref-Grader}). We collected a large set of reference solutions for both the IMO Shortlist and MathArena datasets from the \href{https://artofproblemsolving.com}{AoPS} forum. We use the following workflow that exploits reference solutions to improve the quality and calibration of grading:
 
 \begin{enumerate}
 \item \textbf{Reference Solution Clustering}: The model clusters the reference solutions into groups based on their similarity. 
 \item \textbf{Solution Matching}: The model finds the most similar group of reference solutions to the given solution and uses it as a reference to grade the given solution.
 \item \textbf{Solution Analysis}: The model analyzes the reference solution and breaks it into the main steps based on the "aha moments (main ideas of the solution)" and its substeps.
\item \textbf{Rubric Design}: The model distributes 7 points among the main steps and defines rules on how to allocate points to the substeps. 
\item \textbf{Grading}: The model detects errors in two ways: (1) direct error detection, or (2) contradictions with the reference solution. Contradictions imply the given solution is wrong at that step. Then the model matches the correct and erroroneus parts of the given solution with the rubrics and decides the final grade.
\end{enumerate}

\begin{figure}[h]
   \centering
   \includegraphics[width=0.85\linewidth]{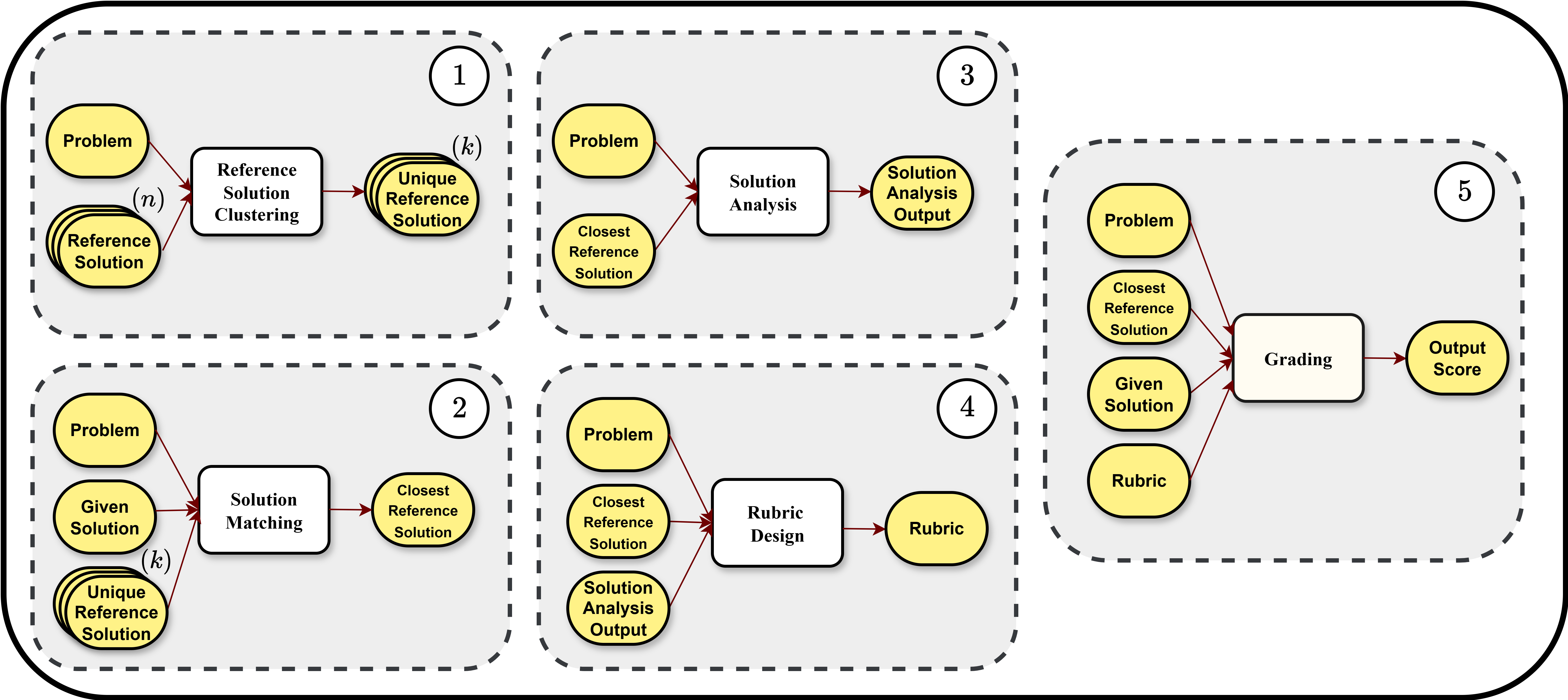}
   \caption{The high-level schema of our multi-stage grading workflow}
   \label{fig:grader-workflow}
\end{figure}

The schema of the workflow is shown in Figure \ref{fig:grader-workflow}. Each of the steps above is a single model call with a specific prompt. Prompts for all steps are provided in Appendix \ref{app:reference-grader}.

\begin{wrapfigure}{r}{0.33\textwidth}
   \centering
   \includegraphics[width=0.33\textwidth]{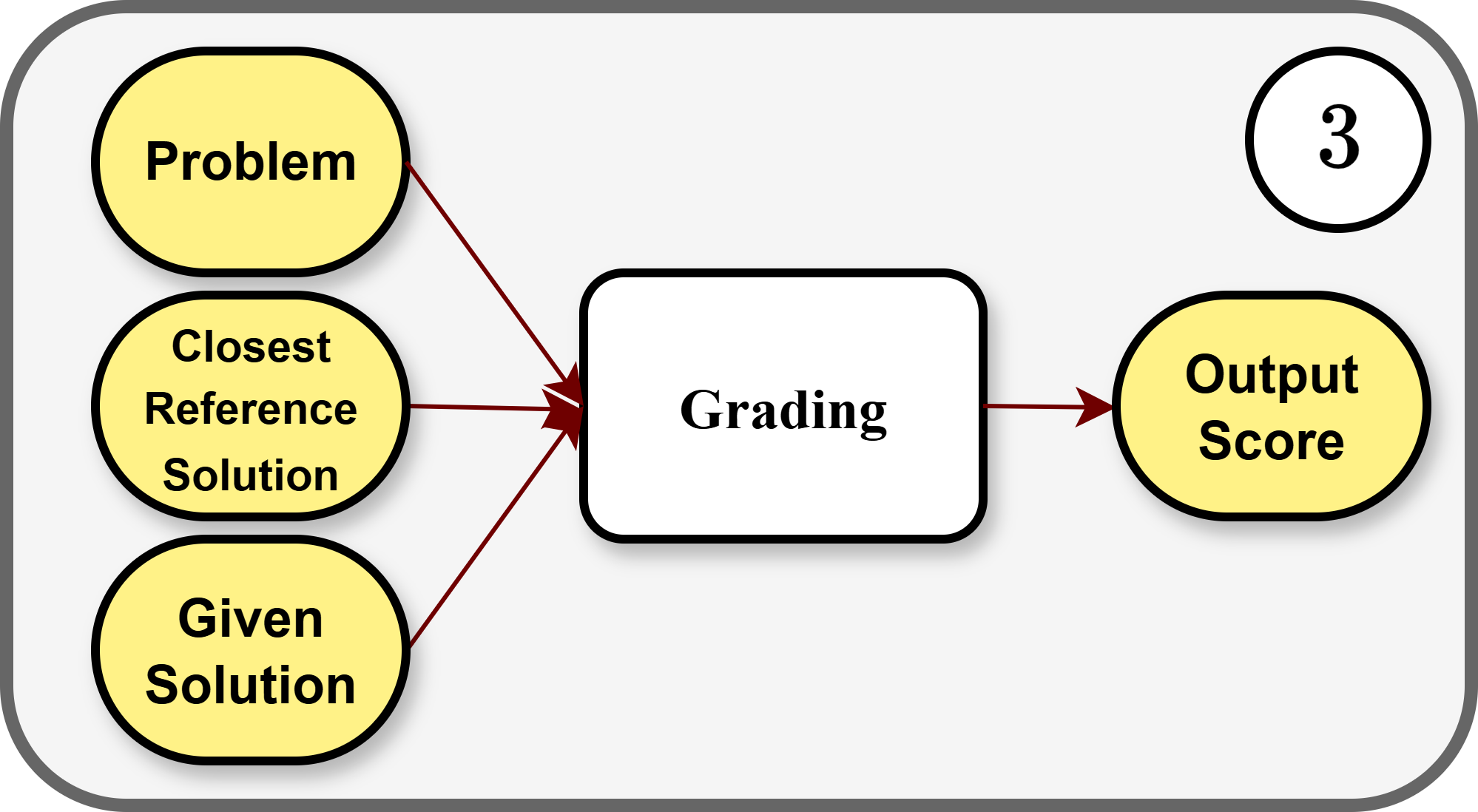}
   \caption{Workflow: reference solution clustering, solution matching, and grading.}
   \label{fig:grader-workflow-ablation}
\end{wrapfigure}

\paragraph{Ablations and settings.} To study the role of each component, we compare the \emph{Single-turn Grader} (one model call without reference solutions), the \emph{5-step Ref-Grader (Plain)} (full workflow with reference solutions, solution analysis, and rubric design), the \emph{5-step Ref-Grader (Approachability)}, which in step 3 computes step-level approachability scores (1-5, measuring how hard a main step is to be chosen) and in step 4 allocates rubric points proportional these scores, the \emph{5-step Ref-Grader (Milestones)}, which in step 4 designs the rubric by milestones reached (milestones denote proving the same or an equivalent intermediate statement as in the reference solution up to a specific step) and the \emph{5-step Ref-Grader (Hybrid)}, which combines the approachability-based analysis with the milestone-based rubric.

We also evaluate the \emph{3-step Ref-Grader (No Rubrics)}, in which step 3 uses a single-turn grading prompt with the reference solution added, without solution analysis and rubric design (Figure~\ref{fig:grader-workflow-ablation} illustrates this variant). This ablation isolates the effect of adding a reference solution and highlights the additional contributions of solution analysis and rubric design.

Tables \ref{tab:multi-turn_matharena} and \ref{tab:multi-turn_imo} summarize the results. Across both datasets, the 5-step Ref-Grader (Approachability) and 5-step Ref-Grader (Milestones) achieve the strongest metrics overall. The 5-step Ref-Grader (Plain) typically ranks third.  The 3-step Ref-Grader (No Rubrics) outperforms the Single-turn Grader on most metrics, indicating that adding a similar reference solution helps even without explicit rubric generation. Overall, these results show that both the reference solution and the rubric contribute to grader performance significantly, and that more careful rubric design brings additional gains. The pearson and rank correlation metrics are shift invariant, as a result, they don't account for the cases where we get the associations right while we are systematically underestimating or overestimating the scores. MAE, RMSE, Off-by-1 and Off-by-2 denote the gaps between real and predicted scores. Finally, QWK and AC2 consider both association and the gap between real and predicted scores.

Interestingly, the 5-step Ref-Grader (Hybrid) performs worse than the other 5-step variants, likely because approachability interferes with milestones: approachability is a property of a reference solution’s step and it assigns a score based on the approach of the reference solution, whereas a milestone can be independent of a specific reference solution, so the two notions are not fully compatible. As a practical note, steps 1 (reference clustering), 3 (solution analysis), and 4 (rubric design) \textbf{can be cached offline}, as they do not depend on the specific given solution. Thus, only steps 2 and 5 need to run online per each given solution. This amortizes the cost of the 5-step workflow.
{\scriptsize \setlength{\tabcolsep}{2pt}\renewcommand{\arraystretch}{0.78}%
\begin{table}[h]
    \centering
    \begin{tabular*}{\linewidth}{@{\extracolsep{\fill}}p{0.36\linewidth}
                    S[table-format=1.2]
                    S[table-format=1.2]
                    S[table-format=1.2]
                    S[table-format=1.2]
                    S[table-format=1.2]
                    S[table-format=1.2]
                    S[table-format=1.2]
                    S[table-format=1.2]@{}}
    \toprule
    \multicolumn{1}{c}{\textbf{Method}} & \multicolumn{1}{c}{\(\mathbf{r\,\uparrow}\)} & \multicolumn{1}{c}{\(\boldsymbol{\rho}\,\uparrow\)} & \multicolumn{1}{c}{\textbf{MAE\,\(\downarrow\)}} & \multicolumn{1}{c}{\textbf{RMSE\,\(\downarrow\)}} & \multicolumn{1}{c}{\textbf{QWK\,\(\uparrow\)}} & \multicolumn{1}{c}{\textbf{Off1\,(\(\uparrow\))}} & \multicolumn{1}{c}{\textbf{Off2\,(\(\uparrow\))}} & \multicolumn{1}{c}{\textbf{AC2\,(\(\uparrow\))}} \\
    \midrule
    Single-turn Grader & 0.66 & 0.63 & 2.32 & 2.75 & 0.36 & 0.32 & 0.49 & 0.36 \\
    3-step Ref-Grader (No Rubrics) & 0.71 & 0.72 & 2.27 & 2.66 & 0.42 & 0.29 & 0.51 & 0.37 \\
    5-step Ref-Grader (Plain) & 0.73 & 0.75 & 1.49 & 2.09 & 0.67 & \textcolor{SecondBest}{0.63} & \textcolor{SecondBest}{0.77} & 0.63 \\
    5-step Ref-Grader (Approachability) & \textbf{0.79} & \textbf{0.77} & \textcolor{SecondBest}{1.33} & \textcolor{SecondBest}{1.98} & \textcolor{SecondBest}{0.72} & \textbf{0.67} & \textbf{0.81} & \textcolor{SecondBest}{0.68} \\
    5-step Ref-Grader (Milestones) & \textcolor{SecondBest}{0.78} & 0.71 & \textbf{1.26} & \textbf{1.89} & \textbf{0.73} & \textcolor{SecondBest}{0.63} & \textbf{0.81} & \textbf{0.72} \\
    5-step Ref-Grader (Hybrid) & 0.76 & \textcolor{SecondBest}{0.76} & 1.50 & 2.12 & 0.68 & 0.61 & 0.74 & 0.63 \\
    \bottomrule
    \end{tabular*}
    \caption{MathArena: Single-turn vs multi-step reference grading.}
    \label{tab:multi-turn_matharena}
\end{table}}

{\scriptsize \setlength{\tabcolsep}{2pt}\renewcommand{\arraystretch}{0.78}%
\begin{table}[h]
    \centering
    \begin{tabular*}{\linewidth}{@{\extracolsep{\fill}}p{0.36\linewidth}
                    S[table-format=1.2]
                    S[table-format=1.2]
                    S[table-format=1.2]
                    S[table-format=1.2]
                    S[table-format=1.2]
                    S[table-format=1.2]
                    S[table-format=1.2]
                    S[table-format=1.2]@{}}
    \toprule
    \multicolumn{1}{c}{\textbf{Method}} & \multicolumn{1}{c}{\(\mathbf{r\,\uparrow}\)} & \multicolumn{1}{c}{\(\boldsymbol{\rho}\,\uparrow\)} & \multicolumn{1}{c}{\textbf{MAE\,\(\downarrow\)}} & \multicolumn{1}{c}{\textbf{RMSE\,\(\downarrow\)}} & \multicolumn{1}{c}{\textbf{QWK\,\(\uparrow\)}} & \multicolumn{1}{c}{\textbf{Off1\,(\(\uparrow\))}} & \multicolumn{1}{c}{\textbf{Off2\,(\(\uparrow\))}} & \multicolumn{1}{c}{\textbf{AC2\,(\(\uparrow\))}} \\
    \midrule
    Single-turn Grader & 0.60 & 0.60 & 1.76 & 2.21 & 0.43 & 0.50 & 0.69 & 0.48 \\
    3-step Ref-Grader (No Rubrics) & 0.70 & 0.71 & 1.58 & 2.04 & 0.55 & 0.51 & 0.78 & 0.57 \\
    5-step Ref-Grader (Plain) & \textbf{0.73} & \textbf{0.74} & 1.26 & 1.83 & 0.70 & 0.66 & \textcolor{SecondBest}{0.83} & 0.75 \\
    5-step Ref-Grader (Approachability) & \textbf{0.73} & \textbf{0.74} & \textbf{1.19} & \textbf{1.75} & \textbf{0.72} & \textbf{0.69} & \textcolor{SecondBest}{0.83} & \textbf{0.77} \\
    5-step Ref-Grader (Milestones) & \textbf{0.73} & \textcolor{SecondBest}{0.72} & \textcolor{SecondBest}{1.20} & \textcolor{SecondBest}{1.80} & \textcolor{SecondBest}{0.71} & \textcolor{SecondBest}{0.68} & \textbf{0.86} & \textbf{0.77} \\
    5-step Ref-Grader (Hybrid) & 0.66 & 0.65 & 1.36 & 1.97 & 0.64 & 0.63 & 0.80 & 0.71 \\
    \bottomrule
    \end{tabular*}
    \caption{IMO Shortlist: Single-turn vs multi-step reference grading.}
    \label{tab:multi-turn_imo}
\end{table}}

\begin{figure*}[h]
    \centering
    \caption*{MathArena}
    \begin{subfigure}[t]{0.31\textwidth}
        \centering
        \includegraphics[width=\linewidth]{Figures/confusion_normalized_gemini_pro_math_arena_absolute_grade_with_tags_math_arena_results.png}
        \caption{Single-turn}
    \end{subfigure}\hfill
    \begin{subfigure}[t]{0.31\textwidth}
        \centering
        \includegraphics[width=\linewidth]{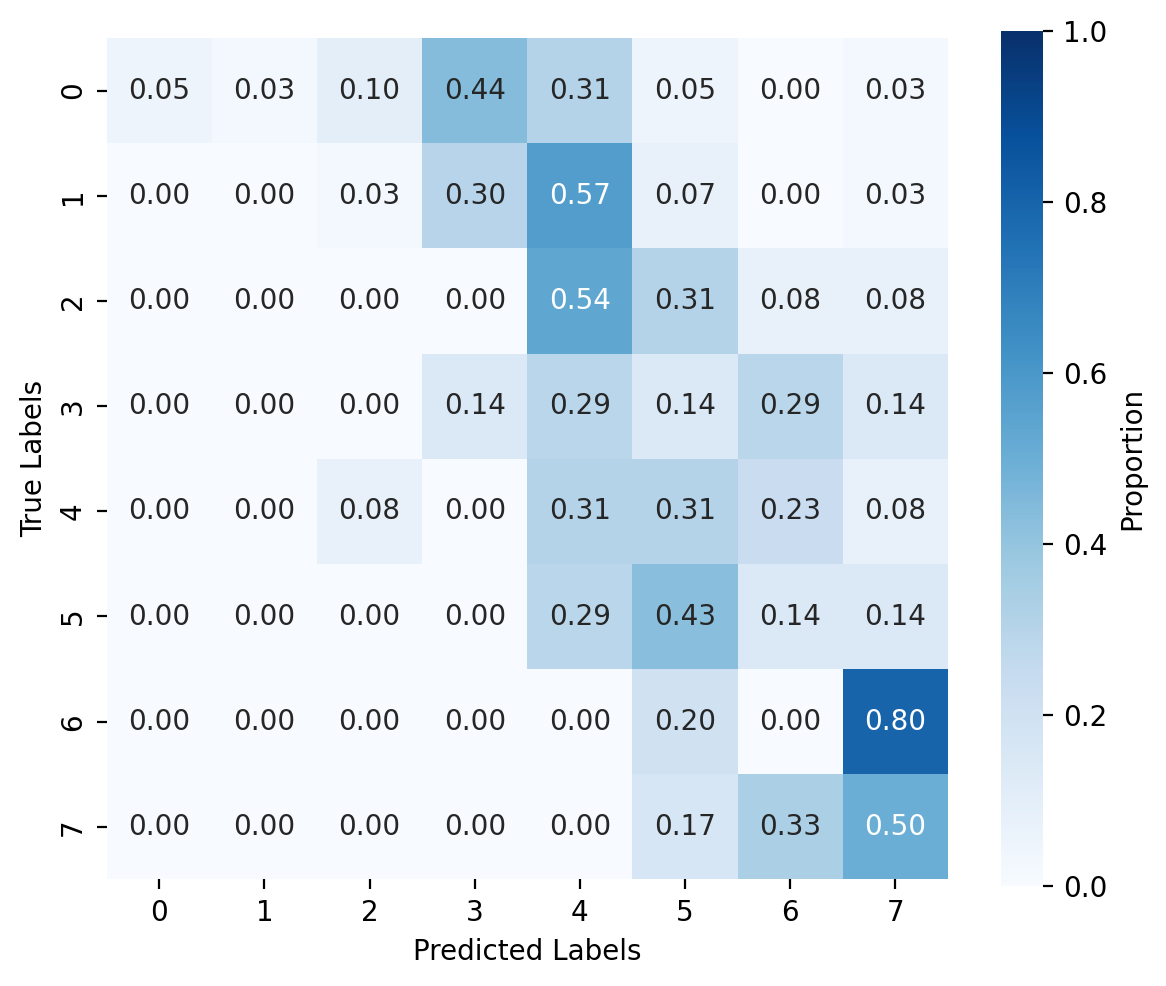}
        \caption{3-step (No Rubrics)}
    \end{subfigure}\hfill
    \begin{subfigure}[t]{0.31\textwidth}
        \centering
        \includegraphics[width=\linewidth]{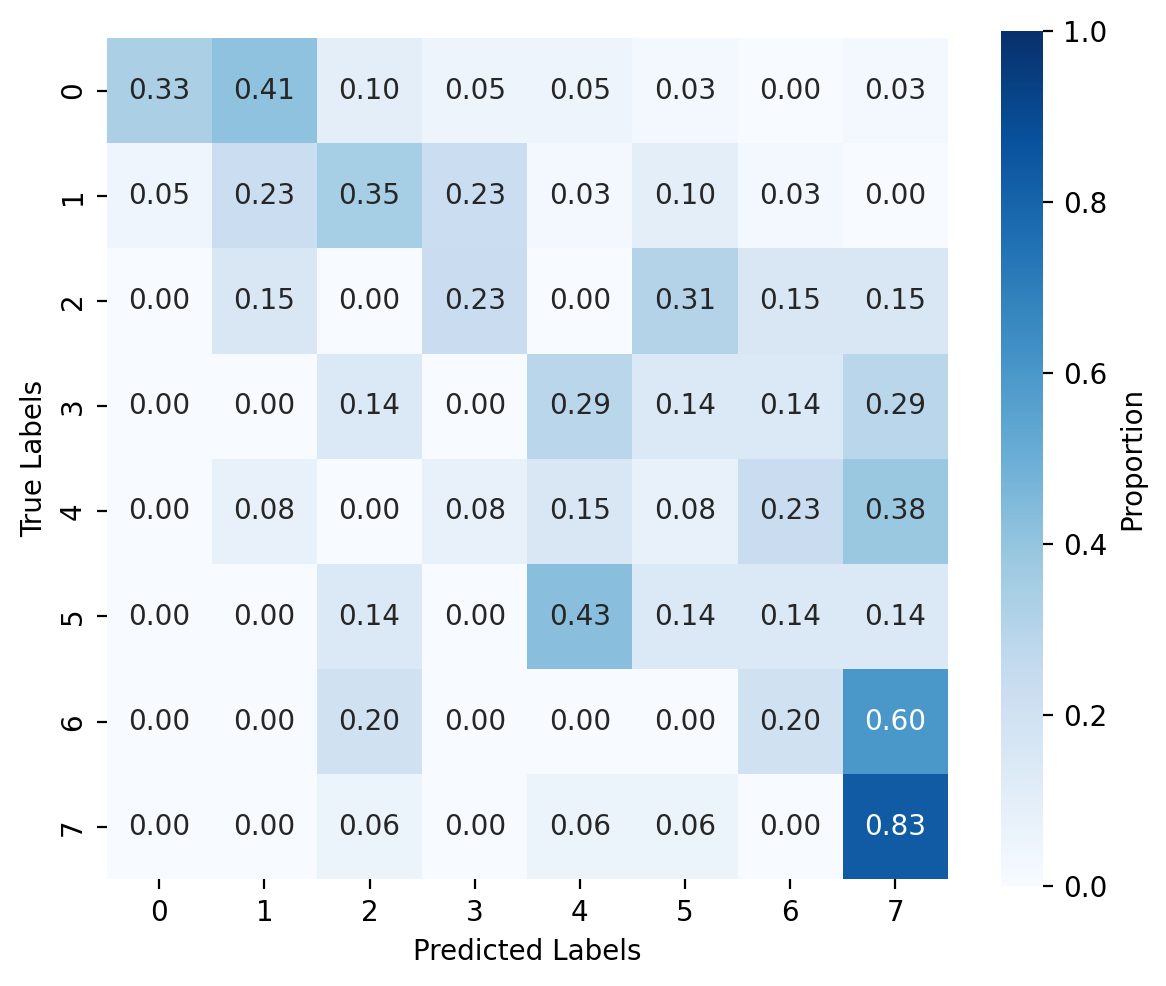}
        \caption{5-step (Plain)}
    \end{subfigure}
    \\[2mm]
    \begin{subfigure}[t]{0.31\textwidth}
        \centering
        \includegraphics[width=\linewidth]{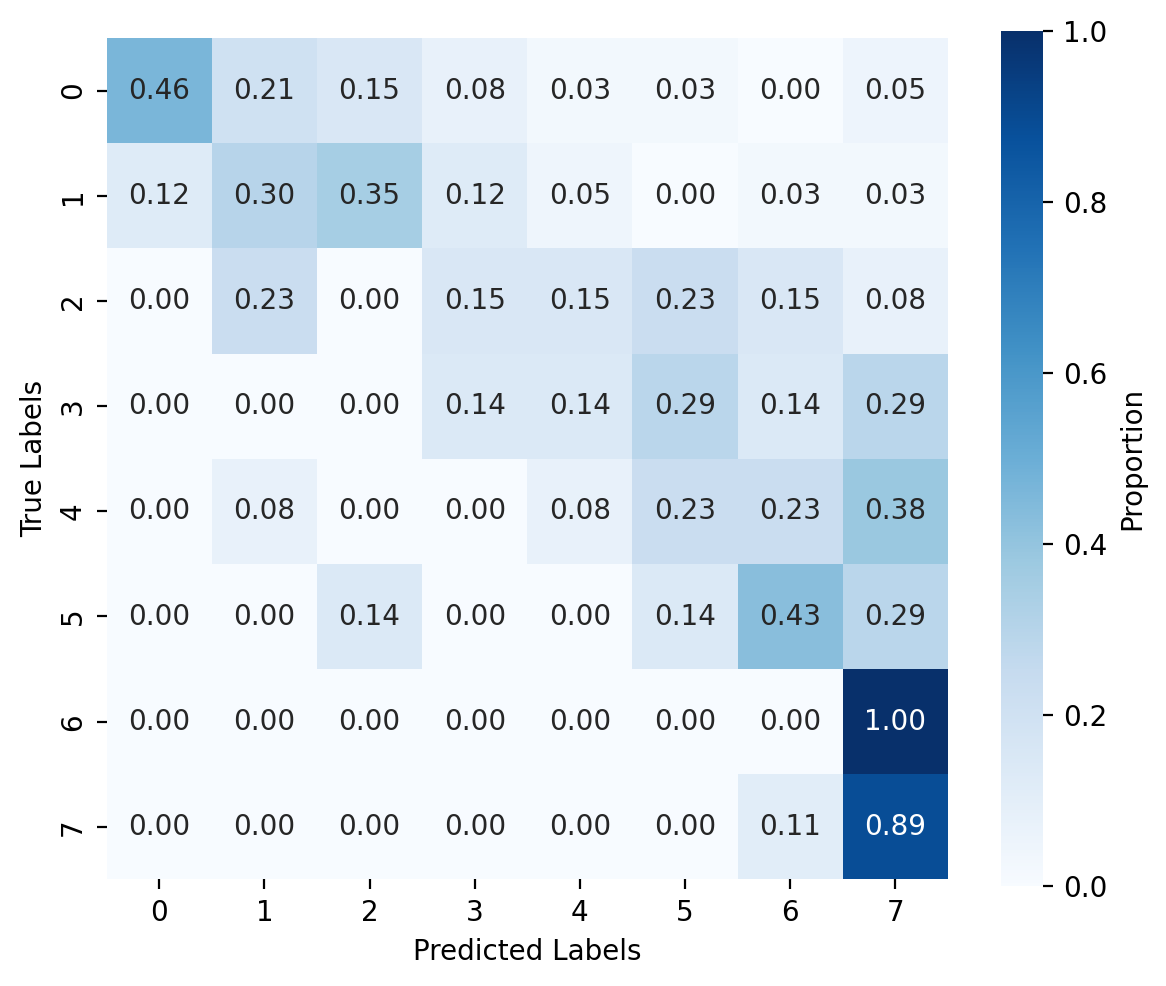}
        \caption{5-step (Approachability)}
    \end{subfigure}\hfill
    \begin{subfigure}[t]{0.31\textwidth}
        \centering
        \includegraphics[width=\linewidth]{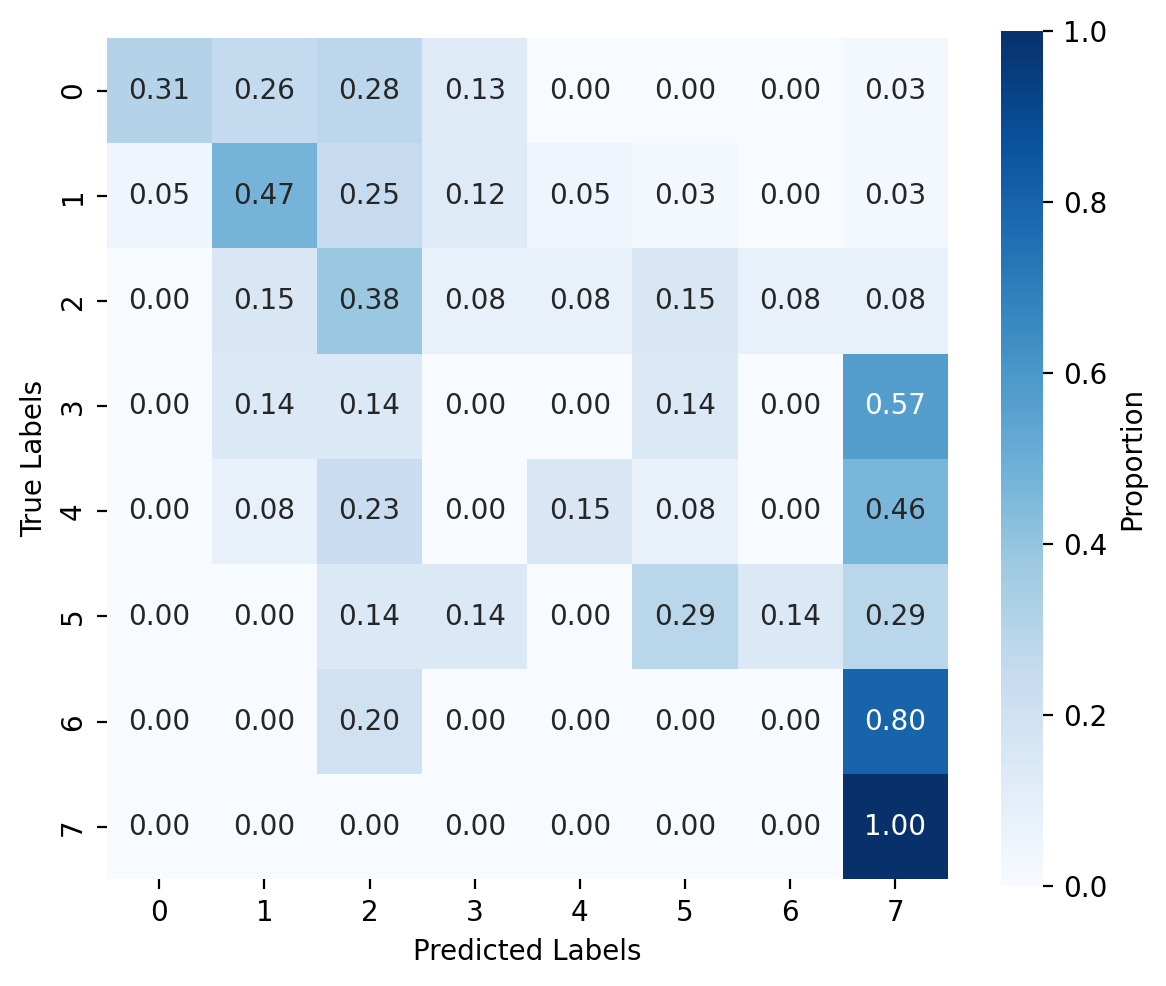}
        \caption{5-step (Milestones)}
    \end{subfigure}\hfill
    \begin{subfigure}[t]{0.31\textwidth}
        \centering
        \includegraphics[width=\linewidth]{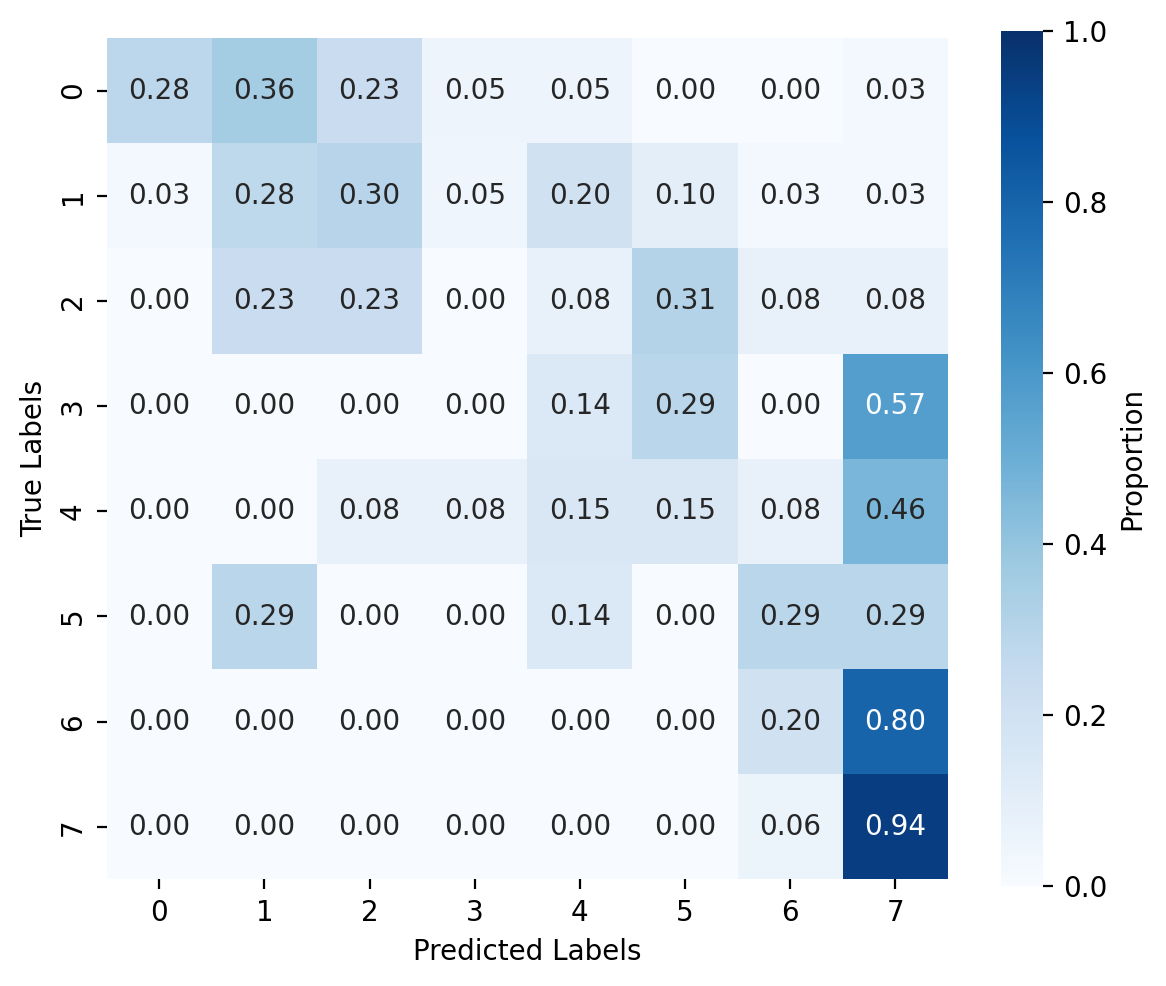}
        \caption{5-step (Hybrid)}
    \end{subfigure}
    \\[2mm]
    \caption*{IMO Shortlist}
    \begin{subfigure}[t]{0.31\textwidth}
        \centering
        \includegraphics[width=\linewidth]{Figures/confusion_normalized_gemini_pro_absolute_grade_no_tags_intermediate_results.png}
        \caption{Single-turn}
    \end{subfigure}\hfill
    \begin{subfigure}[t]{0.31\textwidth}
        \centering
        \includegraphics[width=\linewidth]{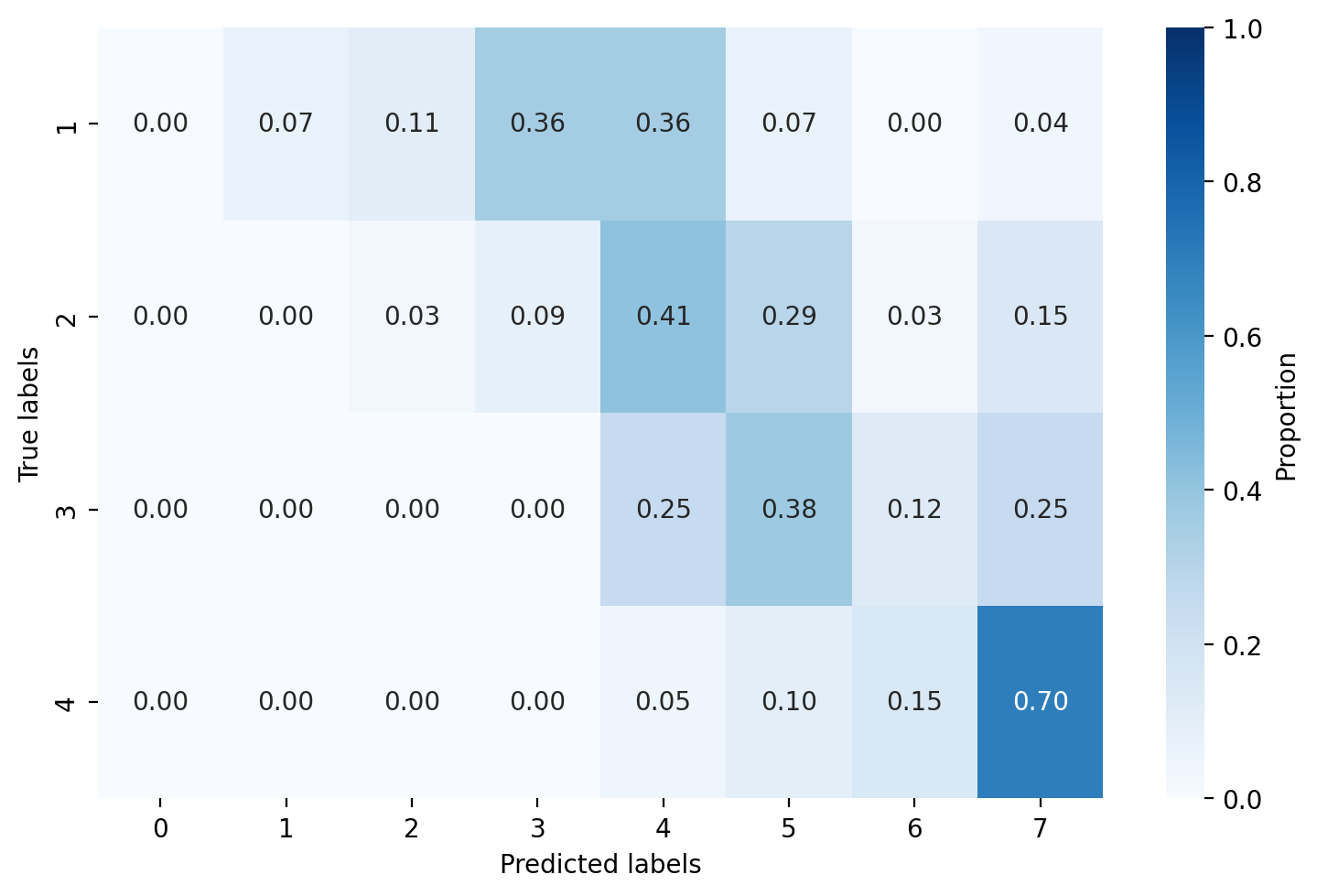}
        \caption{3-step (No Rubrics)}
    \end{subfigure}\hfill
    \begin{subfigure}[t]{0.31\textwidth}
        \centering
        \includegraphics[width=\linewidth]{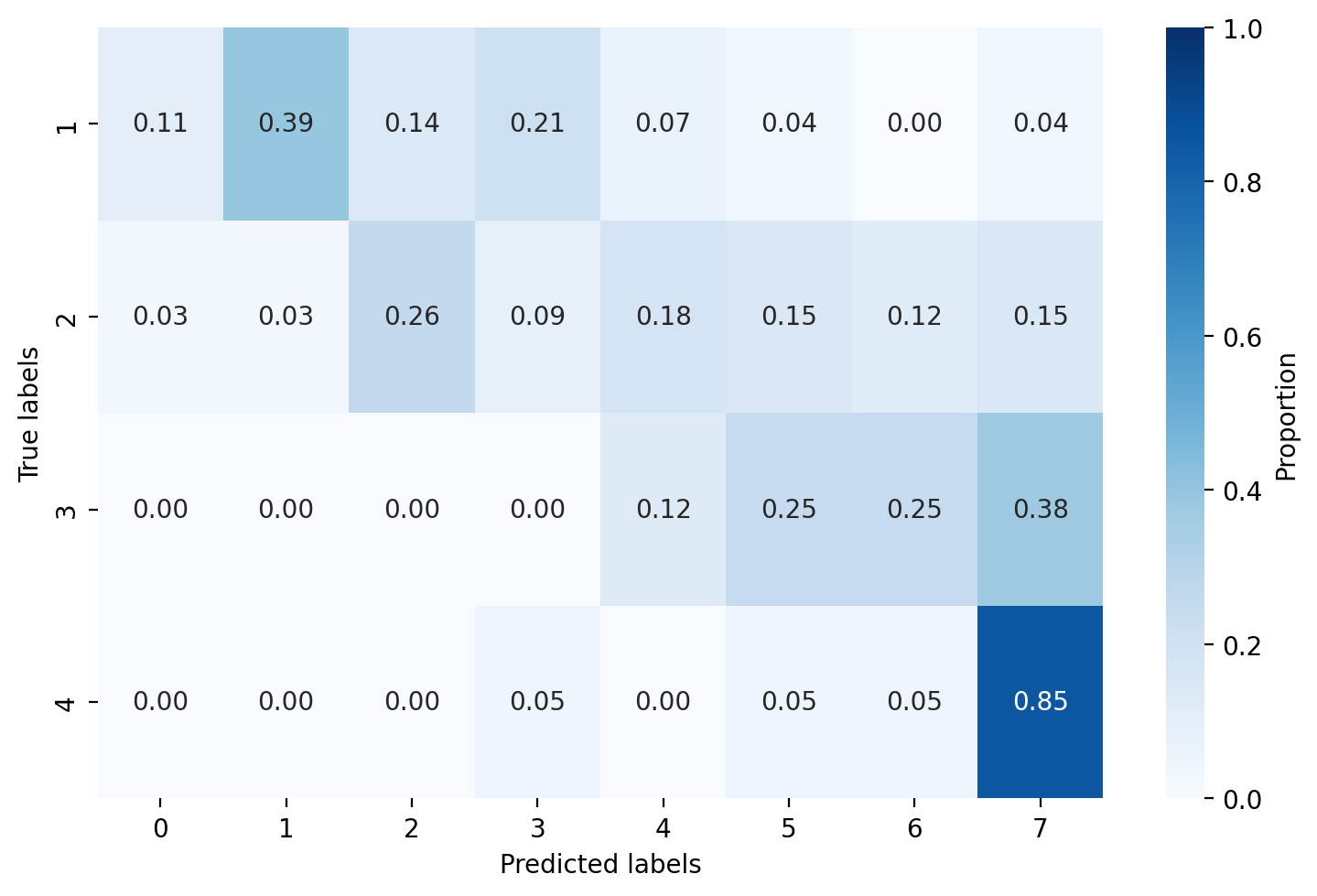}
        \caption{5-step (Plain)}
    \end{subfigure}
    \\[2mm]
    \begin{subfigure}[t]{0.31\textwidth}
        \centering
        \includegraphics[width=\linewidth]{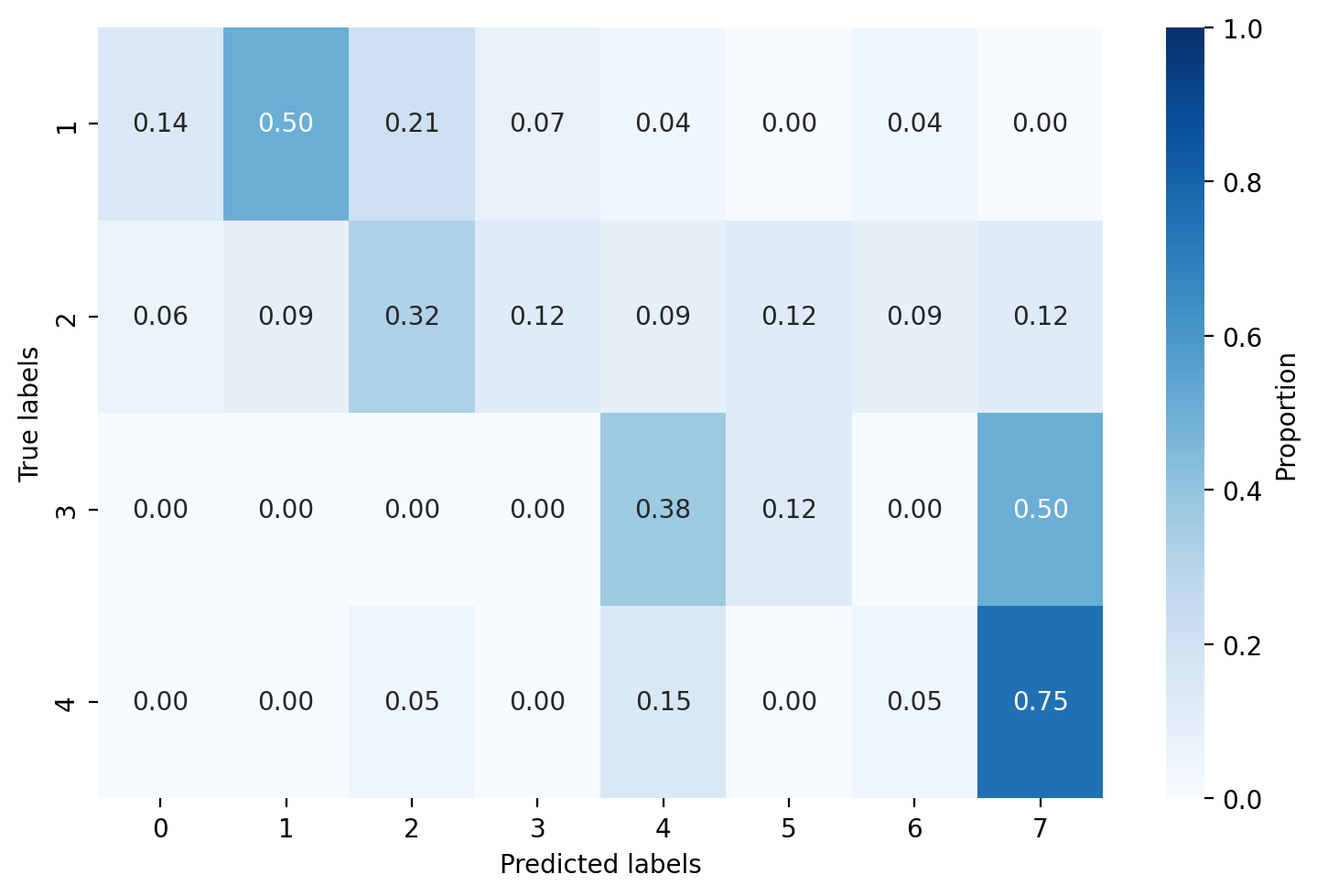}
        \caption{5-step (Approachability)}
    \end{subfigure}\hfill
    \begin{subfigure}[t]{0.31\textwidth}
        \centering
        \includegraphics[width=\linewidth]{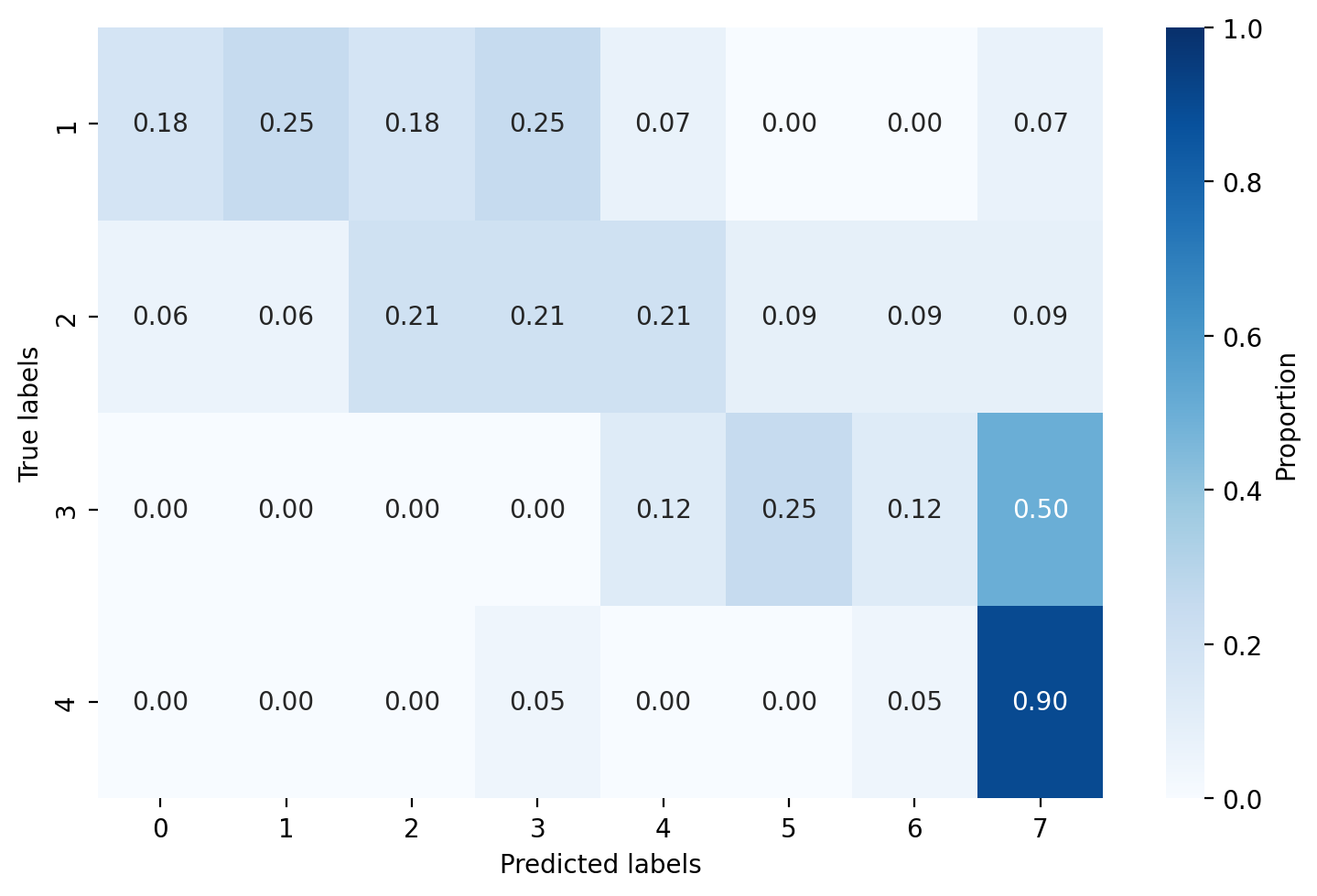}
        \caption{5-step (Milestones)}
    \end{subfigure}\hfill
    \begin{subfigure}[t]{0.31\textwidth}
        \centering
        \includegraphics[width=\linewidth]{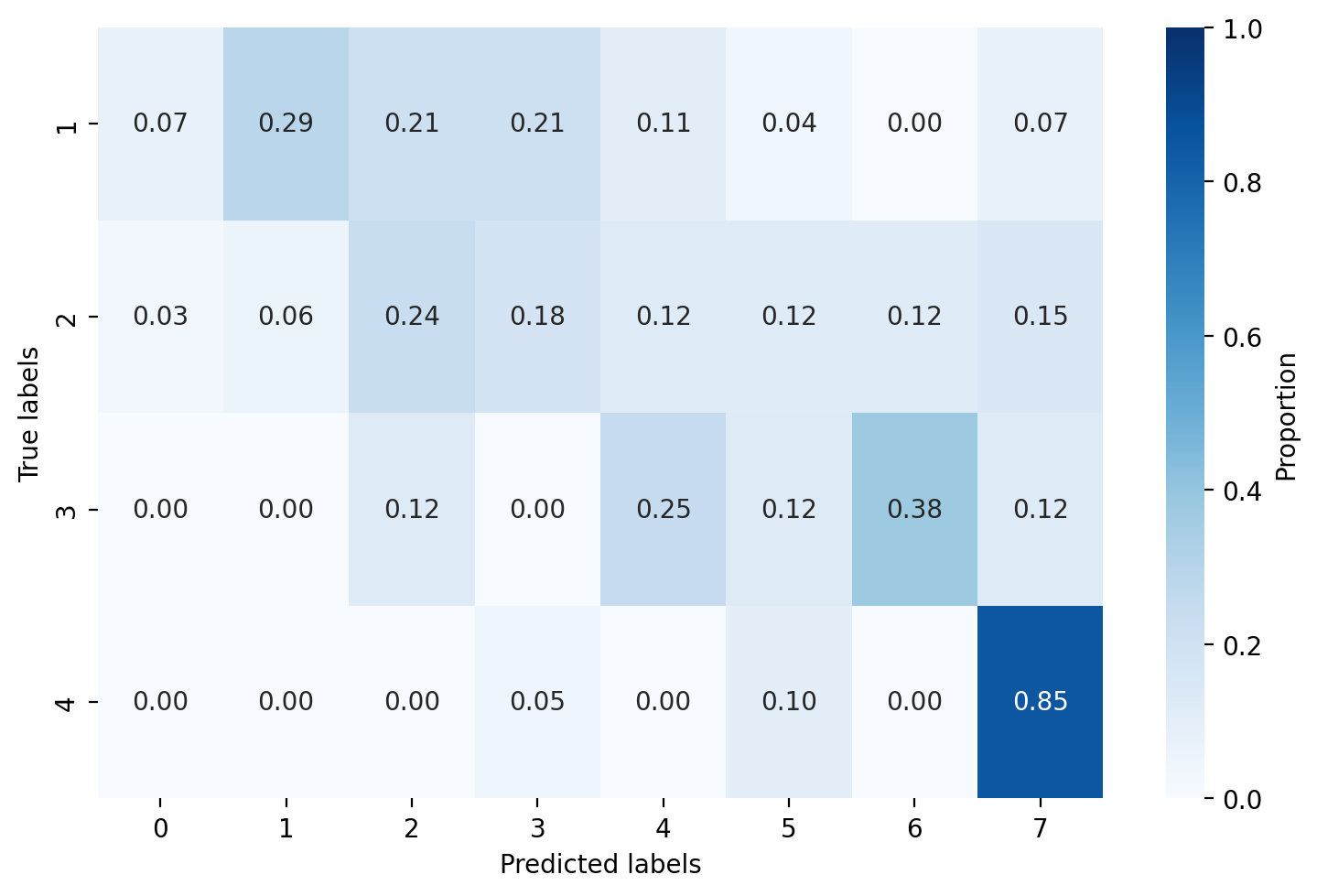}
        \caption{5-step (Hybrid)}
    \end{subfigure}
    \caption{Normalized confusion matrices by each method for MathArena and IMO Shortlist.}
    \label{fig:cm-appendix-grid}
\end{figure*}

\subsection{Are we detecting progress?}
Our main promise is to detect partial progress in incomplete solutions. To make this concrete, we examine the normalized confusion matrices for each method on MathArena and the IMO Shortlist (Figure \ref{fig:cm-appendix-grid}). In an ideal scenario, probability mass should be concentrated along the diagonal. We define partial-progress solutions as those with true score above 0 for MathArena and score label above 1 for the IMO Shortlist. As seen for the single-turn grader earlier, it struggles to produce calibrated scores. By looking at the predicted scores for the solutions with partial progress, several patterns emerge. The single-turn grader almost never produces scores below 3, and using a hard threshold on its output does not reliably separate zero-progress from non-zero-progress solutions. The 3-step RefGrader (No Rubrics) alleviates this to some extent: probability mass shifts toward lower scores for zero-progress solutions and toward higher predicted scores for higher true scores, though calibration issues remain visible in the confusion matrix.

The 5-stage workflows look different, and a diagonal pattern starts to appear, albeit imperfectly. For these workflows, probability mass for low-score solutions shifts toward their true scores, and solutions with true scores of 0 and 1 for MathArena and 1 for the IMO Shortlist can be discriminated with high accuracy by applying a threshold to the predicted output scores.

At the other extreme, one can also ask whether complete solutions can be separated from incomplete ones. For MathArena, all models distinguish the 6 and 7 solutions from the rest quite well. In the context of math olympiads, a 6 solution is a complete solution that contains a minor issue, so discriminating between a 6 and a 7 solution can depend on taste and rubric design rather than an objective mathematical criterion. The same pattern is seen for the IMO Shortlist. All methods assign scores 6 and 7 to solutions with the 4 (correct) label. This further supports our earlier point about the difficulty of separating solutions with progress from those without in comparison to the task of discriminating perfect solutions from incomplete and imperfect ones.

This analysis provides a consistent story for the observed metrics in the previous subsection. The single-turn grader is able to detect perfectly correct solutions versus imperfect ones, but it struggles to detect partial progress. Since all solutions in both datasets are LLM-generated, the solutions are skewed toward zero-progress and completeness; hence, the single-turn grader achieves significant association because it can detect perfectly complete solutions versus imperfect ones. Adding the reference solution and the rubric to the workflow adds the capability of detecting partial progress, which further improves associational and non-associational metrics.

\begin{figure*}[h]
    \centering
    \begin{subfigure}[t]{0.48\textwidth}
        \centering
        \includegraphics[width=\linewidth]{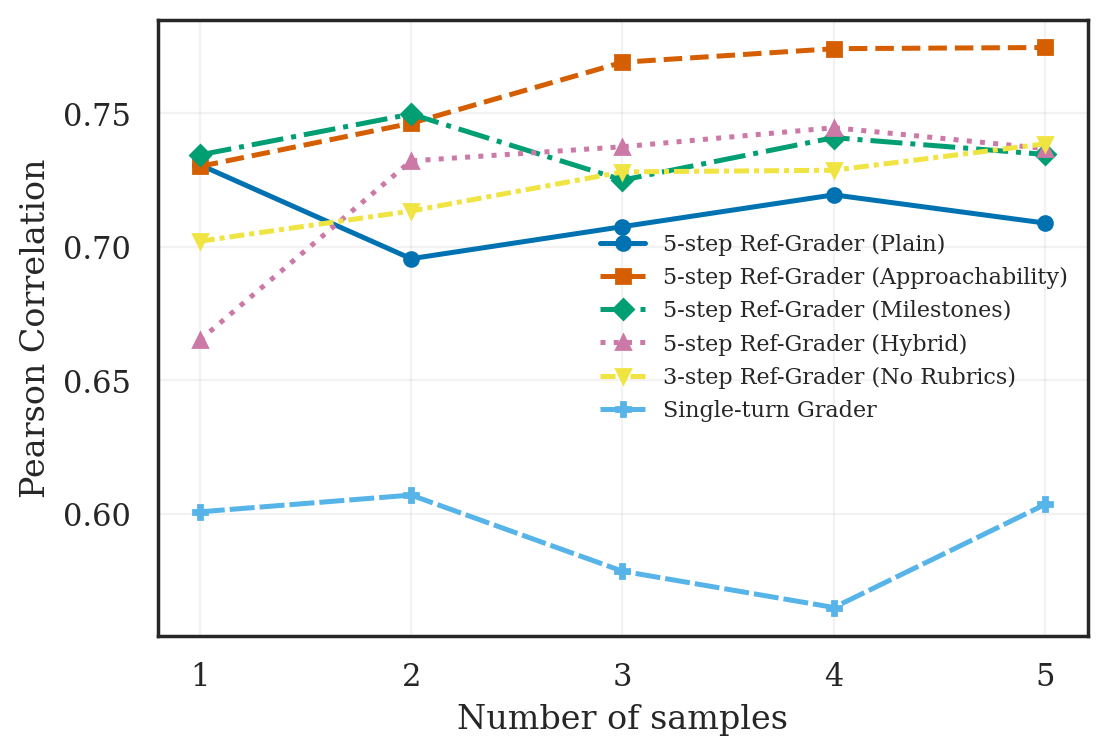}
        \caption{Pearson (\(\uparrow\)).}
    \end{subfigure}\hfill
    \begin{subfigure}[t]{0.48\textwidth}
        \centering
        \includegraphics[width=\linewidth]{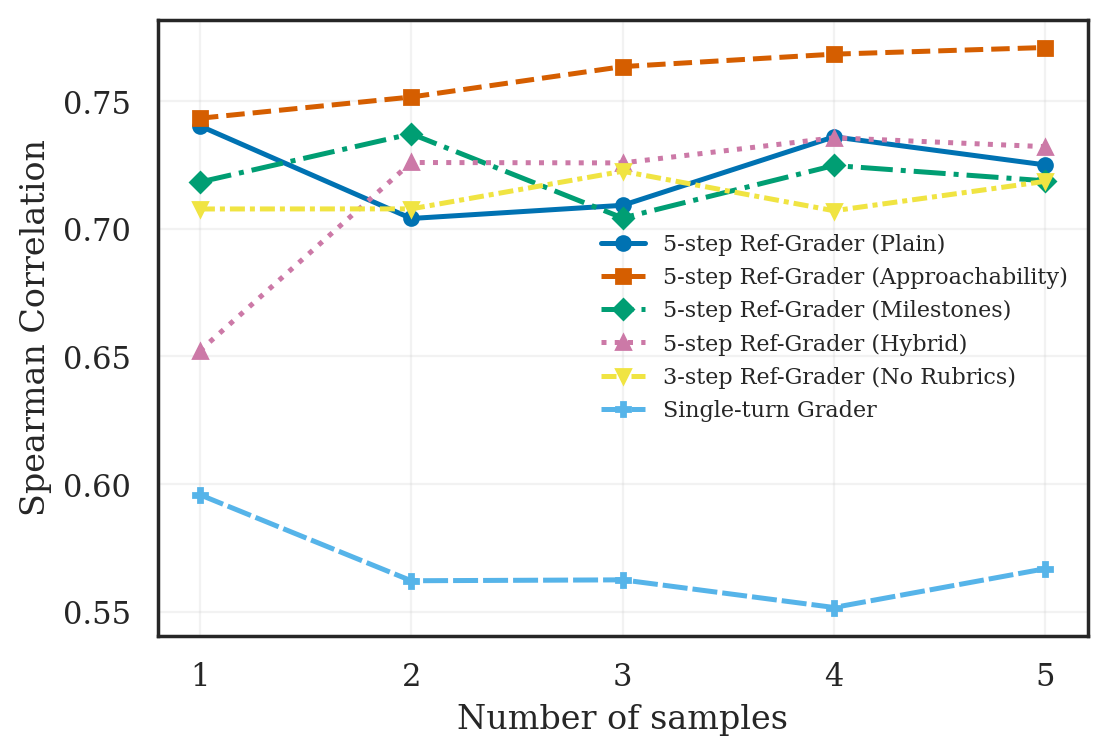}
        \caption{Spearman (\(\uparrow\)).}
    \end{subfigure}
    
    \vspace{2mm}
    \begin{subfigure}[t]{0.48\textwidth}
        \centering
        \includegraphics[width=\linewidth]{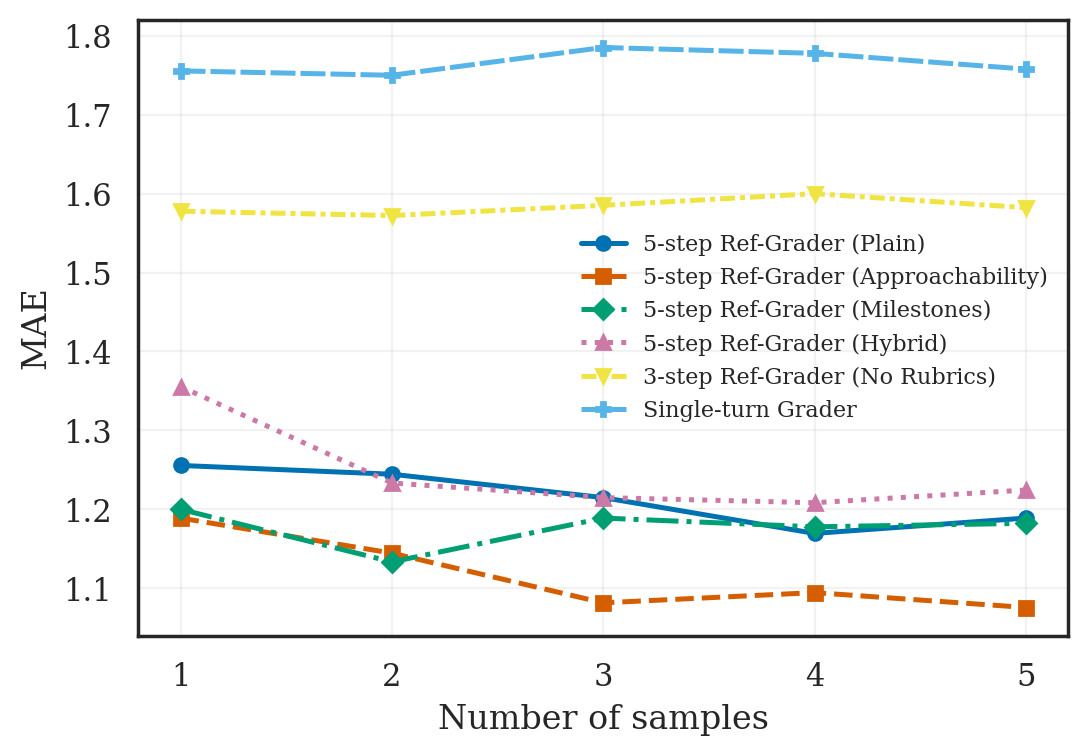}
        \caption{MAE (\(\downarrow\)).}
    \end{subfigure}\hfill
    \begin{subfigure}[t]{0.48\textwidth}
        \centering
        \includegraphics[width=\linewidth]{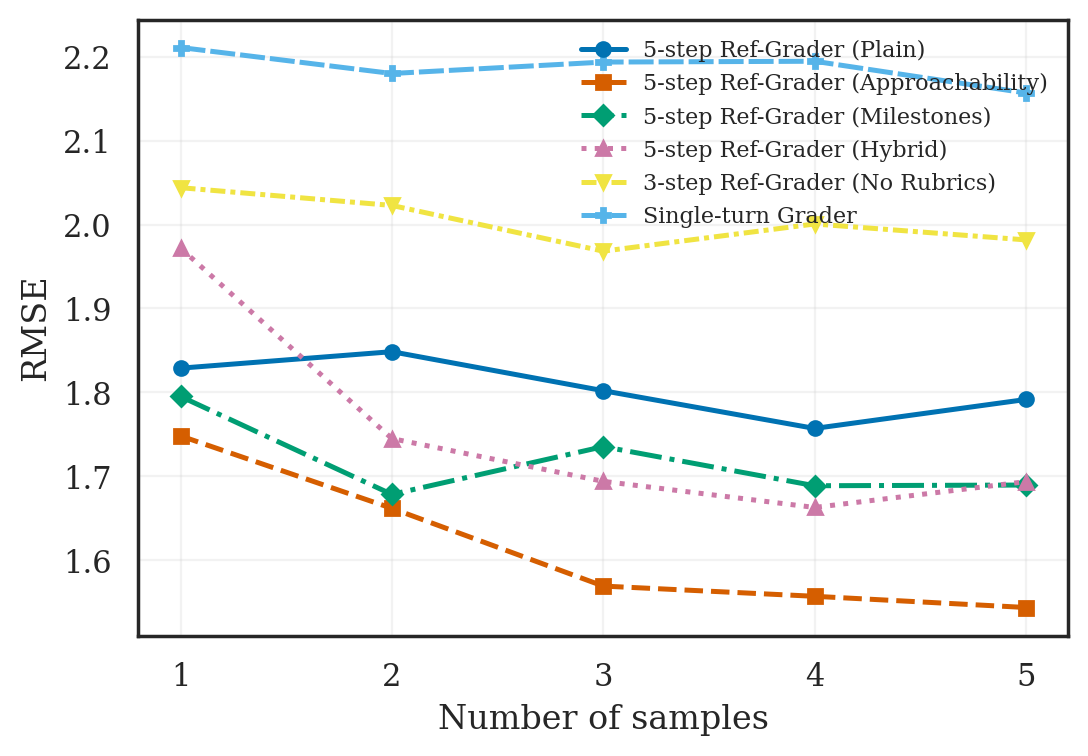}
        \caption{RMSE (\(\downarrow\)).}
    \end{subfigure}
    \caption{Sampling trends for the grader steps across methods for the IMO Shortlist dataset. As we can see, sampling and averaging the grader steps does not consistently add much benefit except for the 5-step Ref-Grader (Approachability).}
    \label{fig:sampling-trends}
\end{figure*}

\subsection{Sampling and Averaging}

 We mentioned that the multi-step grading workflow costs more than the single-turn grading workflow. It is natural to ask whether sampling and averaging the outputs of the single-turn grader explains the gains. Figure \ref{fig:sampling-trends} plots sampling trends for all workflows. With the exception of the 5-step Ref-Grader (Approachability), within-method sampling and averaging yields no consistent performance gains, indicating that improvements are not due to spending more tokens relative to the single-turn grader.

 Interestingly, ensembling across methods can also help. For the IMO Shortlist, averaging predictions from \emph{3-step Ref-Grader (No Rubrics)}, \emph{5-step Ref-Grader (Approachability)}, \emph{5-step Ref-Grader (Plain)}, and \emph{5-step Ref-Grader (Milestones)} achieves Pearson 0.80, Spearman 0.80, MAE 1.11, RMSE 1.52, off-by-one 0.65, and off-by-two 0.82, matching or exceeding the best single-method metrics. A systematic study of ensembling strategies is left for future work.

\section{Conclusion}
We studied proof grading for Olympiad-level mathematics and showed that reference-aided, multi-step workflows substantially improve partial-credit calibration over single-turn graders. Across the IMO Shortlist and MathArena datasets, our 5-step Ref-Grader variants consistently increase agreement with human judges, with approachability-weighted and milestone-based rubrics offering complementary strengths. Ablations indicate that adding a similar reference solution helps even without rubric induction, while sampling/averaging within a method does not explain the gains. 

Beyond evaluation, these workflows support broader uses. First, as LLM-as-a-judge, they provide transparent, step-referenced rationales and more stable partial-credit decisions than rubric-free judging. Second, as a generative reward model for reinforcement learning, the rubric-informed, reference-grounded scoring can shape trajectories toward correct and complete proofs. Third, in education, the same approach can grade student work and surface interpretable feedback on missing steps and error types, provided appropriate reference solutions and guardrails are available. We release data, code, and prompts to facilitate adoption and extensions.

\bibliography{iclr2026_conference}
\bibliographystyle{iclr2026_conference}
\clearpage
\appendix

\section{Prompts}
\label{app:solver-prompt}
\ShowMD{Solver Prompt}{Prompts/solver.MD}

\section{Single Step Grader Prompt}
\label{app:absolute-grader}
\ShowMD{Single-turn Grader Prompt}{Prompts/absolute_grader.MD}

\subsection{Multi-step Grader Workflow Prompts}
\label{app:reference-grader}

\ShowMD{Reference Solution Clustering}{Prompts/similarity_assesment/solution_clustering.MD}

\ShowMD{Solution Matching}{Prompts/similarity_assesment/similarity_assesment.MD}

\ShowMD{Solution Analysis (plain)}{Prompts/solution_analysis/plain.MD}

\ShowMD{Rubric Design (plain)}{Prompts/rubrics_design/plain.MD}

\ShowMD{Grader with Rubrics}{Prompts/relative_grader_with_explicit_error_analysis.MD}

\section*{Ablation Prompts}

\ShowMD{Approachability Based Solution Analysis}{Prompts/solution_analysis/approachability_based.MD}

\ShowMD{Approachability Based Rubric Design}{Prompts/rubrics_design/approachability_based.MD}

\ShowMD{Milestone Based Rubric Design}{Prompts/rubrics_design/milestone_based.MD}

\ShowMD{Hybrid Milestone Based with Approachability Rubrics}{Prompts/rubrics_design/milestone_approachability_hybrid.MD}

\ShowMD{3-Stage Grader Ablation}{Prompts/relative_grader_without_rubrics.MD}

\end{document}